\documentclass{article}
\usepackage{arxiv}

\usepackage{amssymb}
\usepackage{amsfonts}
\usepackage{amsmath}
\usepackage{epsfig}
\usepackage{subfigure}
\usepackage{times}
\usepackage{bm}
\usepackage{multirow}
\usepackage{cite}
\usepackage{caption}
\usepackage{graphicx}
\usepackage{url}
\usepackage[usenames,dvipsnames,svgnames,table]{xcolor}
\usepackage[capitalise]{cleveref}
\usepackage{tikz}
\usepackage{algorithm}
\usepackage{algorithmic}

\usetikzlibrary{bayesnet}
\allowdisplaybreaks[3]
\begin{document}

\title{Probabilistic Classification Vector Machine for Multi-Class Classification}
\date{}

%
%
\author{
	Shengfei Lyu \\
	School of Computer Science and Technique\\
	University of Science and Technique of China\\
	Hefei, Anhui 230027 \\
	\texttt{saintfe@mail.ustc.edu.cn} \\
	\And
	Xing Tian \\
	School of Computer Science and Technique\\
	University of Science and Technique of China\\
	Hefei, Anhui 230027 \\
	\texttt{txing@mail.ustc.edu.cn} \\
	\And
	Yang Li \\
	School of Computer Science and Technique\\
	University of Science and Technique of China\\
	Hefei, Anhui 230027 \\
	\texttt{csly@mail.ustc.edu.cn} \\
	\And
	Bingbing Jiang \\
	School of Computer Science and Technique\\
	University of Science and Technique of China\\
	Hefei, Anhui 230027 \\
	\texttt{jiangbb@mail.ustc.edu.cn} \\
	\And
	Huanhuan Chen \\
	School of Computer Science and Technique\\
	University of Science and Technique of China\\
	Hefei, Anhui 230027 \\
	\texttt{hchen@ustc.edu.cn} \\
}
\maketitle

{\begin{abstract}
	
The probabilistic classification vector machine (PCVM) synthesizes the 
advantages of both the support vector machine and the relevant vector 
machine, delivering a sparse Bayesian solution to classification 
problems. However, the PCVM is currently only applicable to binary cases. 
Extending the PCVM to multi-class cases via heuristic voting strategies 
such as one-vs-rest or one-vs-one often results in a dilemma where 
classifiers make contradictory predictions, and those strategies 
might lose the benefits of probabilistic outputs. To overcome this 
problem, we extend the PCVM and propose a multi-class probabilistic 
classification vector machine (mPCVM). Two learning algorithms, i.e., 
one top-down algorithm and one bottom-up algorithm, have been 
implemented in the mPCVM. The top-down algorithm obtains the maximum a 
posteriori (MAP) point estimates of the parameters based on an 
expectation-maximization algorithm, and the bottom-up algorithm is 
an incremental paradigm by maximizing the marginal likelihood. The 
superior performance of the mPCVMs, especially when the investigated 
problem has a large number of classes, is extensively evaluated on 
synthetic and benchmark data sets.
	
\end{abstract}

}

%
	
\section{Introduction}
\label{introduction}

Classification is one of the fundamental problems in machine learning and 
has been widely studied in various applications.  The basic classification 
predicts whether one thing belongs to a class or not, which is referred to as 
binary  classification. Binary classification is a fundamental problem and 
has been widely studied in numerous well-developed classifiers. Among them, 
the support vector machine (SVM) \cite{Vapnik1998} is arguably the most popular
\cite{Almasi2014}. However, dissatisfaction has been caused for some 
disadvantages \cite{tipping2001sparse, huanhuan_chen_probabilistic_2009} of the SVM, such as i) 
nonprobabilistic outputs, ii) linear correlation between the number of 
support vectors and the size of the training set, which makes the SVM suffer 
when trained with large data sets.

To overcome the above disadvantages of the SVM, the relevance vector machine (RVM)%
\cite{tipping2001sparse} was proposed in a Bayesian automatic relevance 
determination framework \cite{MacKay1992}, \cite{MacKay1992a}. The RVM obtains a 
few original basis functions (called relevance vectors), whose corresponding 
weights are non-zero, by appropriate formulation of hierarchical priors. To  
improve the computational efficiency of the RVM in training, an accelerated 
strategy for the RVM has been developed by means of maximizing the marginal 
likelihood via a principled and efficient sequential addition and deletion 
of candidate basis functions \cite{tipping2003fast}.

Unfortunately, The RVM might not stick to the principle in the SVM that a positive 
sample should have positive weight while a negative sample should have 
negative weight. Consequently, the RVM is unstable and not robust to kernel 
parameters for classification problems. To overcome this issue, Chen \emph{et al.} 
proposed the probabilistic classification vector machine (PCVM) \cite%
{huanhuan_chen_probabilistic_2009} , which guarantees the consistency of numeric signs between weights 
and class labels\footnote{%
For the convenience of exposition, we assume the labels of binary 
classification case are from \{-1, +1\} if not clearly stated. } by adopting a truncated Gaussian prior over 
weights. Similar to the 
accelerated RVM, the efficient probabilistic classification vector machine\cite%
{chen2014efficient}, a fast version of the PCVM, has been proposed.

Like binary classification, multi-class classification is widely used
and worth exploring. Basically, the research\cite%
{rocha2014multiclass} regards multi-class classification as an extension of 
binary classification by using two tricks, i.e., one-vs-rest and one-vs-one. 
The one-vs-rest strategy uses $C-1$ ($C$ is the number of classes) 
classifiers, each of which determines whether a sample belongs to a certain 
class or not. The one-vs-one strategy builds $C(C-1)/2$ classifiers for 
every pair of classes and each sample is classified to the most likely class 
by the majority vote. These two strategies can extend binary classifiers onto 
multi-class cases. For instance, the popular binary classifier SVM has been 
extended to multi-class classification in the well-known toolbox LIBSVM \cite%
{Chang2011LIBSVM} where one-vs-one strategy is adopted. Similarly, the sparse 
Bayesian extreme learning machine (SBELM) \cite{luo2014sparse} constructs a 
sparse version of the Bayesian extreme learning machine by reducing redundant 
hidden neurons and addresses multi-class classification by pairwise 
coupling (another name for the one-vs-one strategy). 

However, the two strategies both suffer from the problem of ambiguous regions 
\cite{BishopPRML, jiang2017scalable}, where classifiers make contradictory predictions. In 
addition, both strategies could not directly produce probabilistic outputs 
over classes despite the fact that extra work can assist to enable 
probability. For example, for the one-vs-one strategy, additional 
post-processing, e.g., solving a linear equality-constrained convex 
quadratic programming problem\cite{luo2014sparse} produces 
probabilistic outputs. Weston \emph{et. al.} pointed out that a more natural 
way to solve multi-class problems was to construct a decision function by 
considering all classes simultaneously\cite{weston1999support}. Based on
this idea, many works have studied the multi-class problem. A very simple straightforward 
algorithm is the multinomial logistic regression (MLR)\cite%
{friedman2000additive}. Similarly, based on the least squares regression 
(LSR), the discriminative LSR (DLSR) \cite{xiang2012discriminative} is 
proposed to solve classification problems by introducing a technique called $%
\varepsilon$-dragging and translates the one-vs-rest training rule to 
multi-class classification.

Unlike the binary nature of the SVM, the RVM could be extended to solve multi-class 
classification problems without the help of those two above strategies.  
The multi-class relevance vector machine (mRVM) \cite{damoulas_inferring_2008} 
employs multinomial probit  likelihood \cite{albert1993bayesian} by 
calculating  regressors for all classes.  Two versions of the mRVM (the mRVM$_1$ and the
mRVM$_2$) have been implemented in different ways. While the mRVM$_2$  employs a 
flat prior to the hyper-parameters that control the sparsity of the 
resulting model, the mRVM$_1$ is a multi-class extension of maximizing the marginal 
likelihood procedure in \cite{tipping2003fast}.

However, the mRVMs still do not ensure the consistency of numeric signs between 
weights and class labels. Therefore, the mRVMs could still be unstable 
and not robust to kernel parameters. To ensure the consistency in multi-class 
cases, a multi-class classification principle has been defined in 
\cref{specification}. 

To relieve the drawbacks of the mRVMs, we extend the applicability of the PCVM and 
propose a multi-class probabilistic classification vector machine (mPCVM).
The mPCVM introduces two types of truncated Gaussian priors over weights for 
training samples. The two types of priors depend on whether 
training samples belong to a given class or not. By the priors, 
the multi-class classification principle has been implemented in the mPCVM. 

In our paper, two learning algorithms have been investigated, i.e., the
top-down algorithm mPCVM$_1$ and the bottom-up algorithm mPCVM$_2$, an 
online incremental version of maximizing the marginal likelihood.

The main contributions of this paper can be summarized as follows:

\begin{enumerate}
\item A multi-class version of the PCVM has been proposed for multi-class 
classification;
	
\item Compared with the SVM, the mPCVM directly produces probabilistic outputs for 
all classes without any post-processing step;

\item A multi-class classification principle is proposed for the mPCVM. It states that 
weights should be consistent with class labels in multi-class cases;

\item Due to the sparseness-encouraging prior, the generated model is 
sparse and its computational complexity in the test stage has been greatly 
reduced.

\end{enumerate}

The rest of this paper is organized as follows. The preamble and related works 
of the PCVM in \cref{specification} are followed by an introduction of the prior 
knowledge on weights in \cref{prior}. \cref{em} presents the detailed 
expectation-maximization (EM) procedures for the mPCVM$_1$. Then an incremental 
learning algorithm mPCVM$_2$ is introduced in \cref{mpcvm2}. The
experimental results and analyses are given in \cref{exp}. Finally, %
\cref{conclude} concludes our paper and presents future work.

\section{Multi-class Probabilistic Classification Vector Machine}

\label{mpcvm}

This section defines the mathematical notations in the beginning. Scalars are 
denoted by lower case letters, vectors by bold lower case letters, and matrices 
by bold upper case letters. For a specific matrix $\bm{Z}$ of size $N \times 
C$, we use the symbols $\bm{z}_n$ and  $\bm{z}_c$ to 
denote the $n$-th column and the $c$-th row of $\bm{Z}$, respectively. A scalar $%
z_{nc}$ is used to denote the $(n,c)$-th element of $\bm{Z}$.

Let $D = {\{\bm{x}_n,t_n\}}_{n=1}^N$ be a training set of $N$ samples, 
where the vertical vector $\bm{x}_n \in {\mathbb{R}}^d$ denotes a data point,  $t_n \in
\{1,2,\cdots,C\}$ denotes the label for the $ n $-th sample, and $C$ denotes 
the number of classes. The multi-class classification learning objective is to 
learn a classifier $f(\bm{x})$ that takes a vector $\bm{x}$ as input and 
assigns the correct class label to it. A general form of $f(\bm{x})$ is 
given as 
\begin{equation}
f(\bm{x}; \bm{w},b) = \sum_{m=1}^{M}{ \phi_m(\bm{x}) w_m} + b \label{f_rvm},
\end{equation}
where the weight vector $\bm{w} = (w_1,w_2, \cdots, w_M)^{\mathrm T}$ and the bias 
term  $b$  are  parameters of the model. $\bm{\phi}(\cdot) = {(\phi_1(\cdot
	), \phi_2(\cdot), \cdots, \phi_M(\cdot))}^{\mathrm T}$ is a fixed 
nonlinear basis function vector, which maps a data point $\bm{x}$ to a 
feature vector with $ M $ dimensions. 

\subsection{Model Preparation}
\label{specification}

The formulation of the PCVM with binary classification will be given as follows:

\begin{equation}
h(\bm{x}; \bm{w},b) = \Psi (\bm \phi(\bm{x})^{\mathrm T}\bm{w} + b),
\label{pcvm_func}
\end{equation}
where $\Psi(\cdot)$ is the Gaussian cumulative distribution function, and $%
\bm{\phi}(\bm{x})  = [\phi_1(\bm{x}), \phi_2(\bm{x}), ..., \phi_N(\bm{x}%
)]^{\mathrm T}$ the basis function. If ${\bm \phi}(\bm x)^{\rm T}{%
\bm w} + b$ is greater than 0, the datum is more likely to be in class $2$ 
than class $1$ \footnote{%
We use class \{1, 2\} here instead of \{-1, +1\} to be compatible 
with multi-class cases. }. The PCVM makes use of a truncated Gaussian 
prior to constrain weights of basis functions of the class 1 to be 
nonpositive and weights of  basis functions of the class 2 to be 
nonnegative, in order to be consistent with the SVM \cite{jiang2019probabilistic}.

For multi-class cases, we assign a set of 
independent weights to each class and extend the idea of the PCVM such that  weights of basis functions of the class $c \in 
\{1,2,..., C\} $ in the $c$-th weight column 
vector (denoted as $\bm{w}_c$) are restricted as nonnegative for training samples in the class $c$ and 
nonpositive otherwise.
The definition of the multi-class 
classification principle for the mPCVM is given as below.

\noindent \textbf{Definition 1.} \textit{
Given multi-class data $D = {\{\bm{x}_n,t_n\}}_{n=1}^N$ with the class 
labels $(1,2,..., C)$. The \textbf{multi-class classification principle for the mPCVM} is that the weight ${w}_{nc}$ of a datum 
$(\bm{x}_n,t_n)$ is consistent with the class labels if 
\begin{equation} \label{mc_principle}
\begin{cases}
w_{nc}  \geq 0 & \qquad \text{if} \quad t_n = c \\
w_{nc}  \leq 0 & \qquad \text{if} \quad t_n \neq c
\end{cases}.
\end{equation}
}

With this principle, weights are consistent with class labels like binary 
classification. Further, a class-based potential $y_{nc}$ for a training datum 
($\bm{x}_n, t_n $) is defined as 
\begin{equation}
\label{mpcvm_2} y_{nc} \triangleq
\bm{\phi}(\bm{x}_n)^{\rm T}\bm{w}_c + b_c,
\end{equation}%
where ${b}_c$ is the bias for the $c$-th class. Subsequently, for data $(\bm{X},%
\bm{t})$, where $\bm{X} = (\bm{x}_1,\bm{x}_2,...,\bm{x}_N)$ and $\bm{t} = (t_1,t_2,...,t_N)^{\mathrm T}$, we have 
\begin{equation}
\bm{Y} = \bm{\Phi}\bm{W} + \bm{1}\bm{b}^{\mathrm T} \label{Y},
\end{equation}%
where $\bm{W} = (\bm{w_1},\bm{w_2},...,\bm{w_C})$,  $\bm{b} %
= (b_1,b_2,...,b_C)^{\mathrm T}$, $ \bm{\Phi} = (\bm{\phi}(\bm{x}_1),%
\bm{\phi}(\bm{x}_2), ...,  \bm{\phi}(\bm{x}_N))^{\rm T}$, and $\bm{1}  $ 
is an all-1 vertical vector.

For any $\bm{x}_n$, the predicted class is determined by  
$\mathop{\arg\max}_{c}\{y_{nc}\}$. Following the procedure in \cite%
{figueiredo_adaptive_2003}, the term $y_{nc}$ in \cref{mpcvm_2} is 
assumed to be coupled with an auxiliary random variable, which is 
denoted as $z_{nc}$
\begin{equation} \label{hidden_var} 
z_{nc} \triangleq y_{nc} + \varepsilon_{nc} = {\bm{\phi}}(\bm{x}_n)^{\mathrm T}{%
\bm{w}_c} + b_c + \varepsilon_{nc},
\end{equation}
where $\varepsilon_{nc}$ obeys a standard normal distribution (i.e., $\varepsilon_{nc} 
\sim \mathcal{N}(0,1)$).  The joint distribution of $\bm{Z}$  is given as 
\begin{align} \label{Z}
p(\bm{Z}|\bm{W},\bm{b}) = (2 \pi) ^ {\frac{-N \times C}{2}} \exp
\left\{-\frac{1}{2} \sum\limits_{c=1}^{C} \| \bm{z_c}-\bm{y_c} \|^2 \right 
\}.
\end{align}

The connection of noisy term $\bm{z_n}$ to the target $t_n$ as 
defined 
in multinomial probit regression \cite{albert1993bayesian} is formulated as  $%
t_n = i$ if $ z_{ni} > z_{nj}, \forall j \neq i $. Sequentially, the joint 
probability between $t_n = i$ and $\bm{z_n}$ is
\begin{flalign}
\label{p(yz,wb)}
\hspace*{-0.35cm} 
\begin{split}
p(t_n&= i,\bm{z_n}|\bm{W},\bm{b}) = \\
& \delta(z_{ni} > z_{nj}, \forall j \neq i) \prod\limits_{c=1}^{C} { \mathcal{N}(z_{nc}|\bm{\phi}}(x_n)^{\rm T}{\bm{w}}_c + b_c,1),
\end{split}
\end{flalign} where $\delta(\cdot)$ is the indicator function. And we have
\begin{align}
\hspace*{-0.3cm} 
\begin{split}
 p(\bm{t},\bm{Z}|\bm{W},&\bm{b}) = \\
&  \prod\limits_{n=1}^{N} \delta(z_{nt_n} > z_{nj}, \forall j \neq t_n) 
\prod\limits_{c=1}^{C} { \mathcal{N}(z_{nc}|y_{nc},1) }.
\end{split}
\end{align}

By marginalizing the noisy potential $\bm{z_n}$, the multinomial 
probit is obtained as  (more details in Appendix \ref{mul_pro})
\begin{align}
\hspace*{-0.3cm} 
\begin{split}
 p(t_n= &i | \bm{W},\bm{b}) \\
& = \int \delta (z_{ni} > z_{nj}, \forall j \neq i) \prod\limits_{c=1}^{C} {%
\mathcal{N}(z_{nc}|y_{nc},1)} d \bm{z_n} \\
& = \mathbb{E}_{\varepsilon_{ni}} \left[  \prod_{j \neq i}\Psi(\varepsilon_{ni} + y_{ni} 
- y_{nj}) \right].
\end{split}
\label{p(tn,i)}
\end{align} 

And we have
\begin{align}
p(\bm{t}|\bm{W},\bm{b})  = \prod \limits_{n = 1}^{N}  \mathbb{E}_{\varepsilon_{nt_n} }
\left[  \prod_{j \neq t_n}\Psi(\varepsilon_{nt_n}  + y_{nt_n} - y_{nj}) \right].
\label{p(t)}
\end{align}

\subsection{Priors over weights}

\label{prior}

\begin{figure}[t]
	\centering\includegraphics[width=2in]{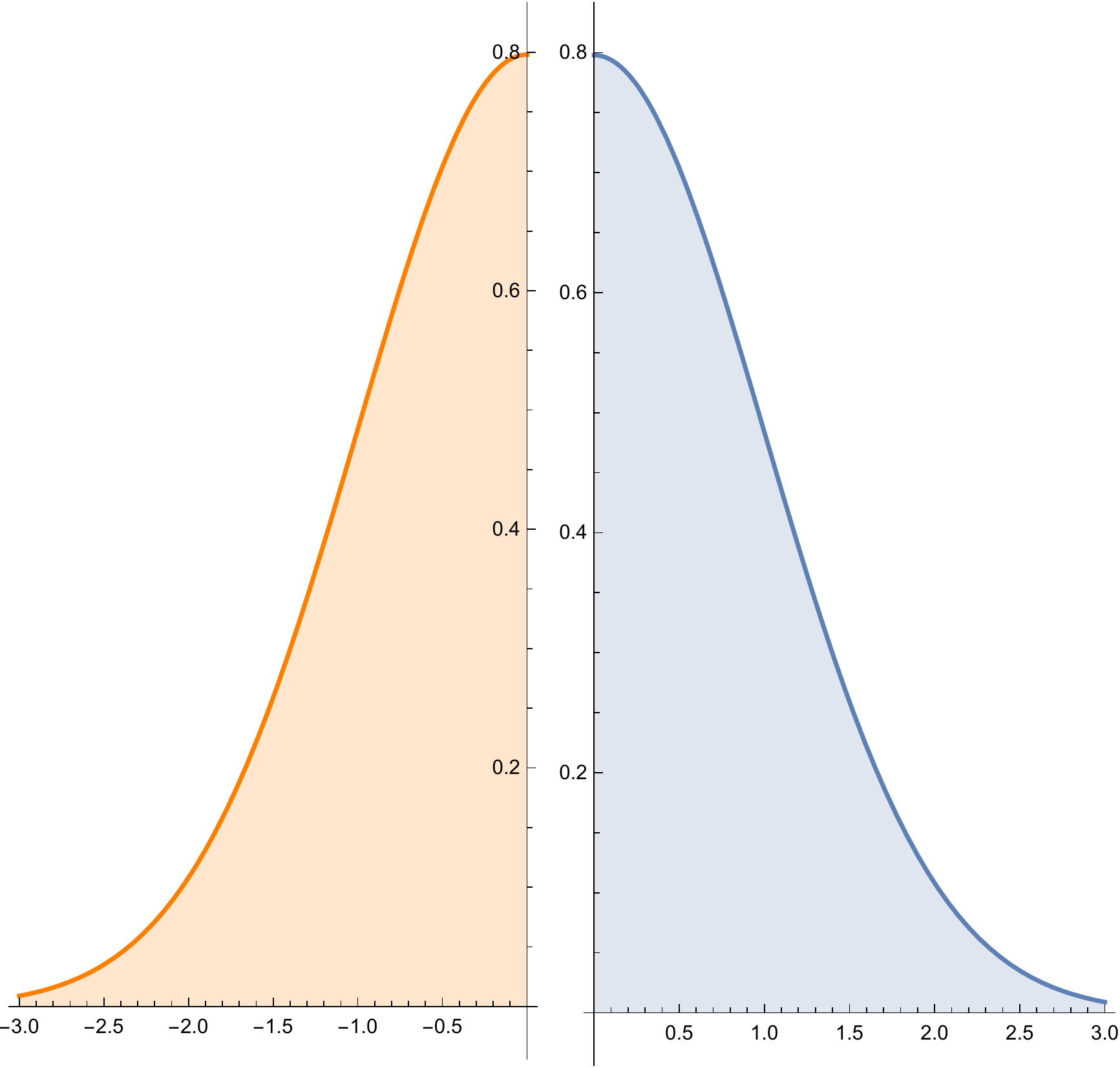} \caption{The truncated
	Gaussian prior over weight $w_{nc}$. Left: when $f_{nc}=-1$,
	$p(w\bm{|}\protect\alpha)$ is a nonpositive, right-truncated
	Gaussian prior. Right: when $f_{nc}=+1$, $p(w\bm{|} \protect\alpha)$
	is a nonnegative, left-truncated Gaussian prior. } \label{truncated_figure}
\end{figure}

As discussed in \cref{specification}, for the sake of ensuring the multi-class 
classification principle, the left-truncated Gaussian prior \cite{chen2009predictive} and the 
right-truncated Gaussian prior over weights are chosen. Their distributions 
are formulated in \cref{truncatd_gaussian} and illustrated in %
\cref{truncated_figure}
\begin{equation}\label{truncatd_gaussian}
\mathcal{N}_t(w_{nc}|0,{\alpha}_{nc}^{-1}) = 2\mathcal{N}(w_{nc}|0,{\alpha}%
_{nc}^{-1})\delta({f_{nc}w_{nc}}  > 0),
\end{equation} 
where ${\alpha}_{nc}$ is the inverse variance and $f_{nc} = 1$ if $ t_n = c $ 
otherwise $ -1$.

 Then, the prior distribution over the weight $\bm{W}$ is 
\begin{equation}\label{w_prior}
 p(\bm{W}|\bm{A}) = \prod_{n=1}^{N}\prod_{n=1}^{C}{\mathcal{N}_t(w_{nc}|0,{%
 \alpha}_{nc}^{-1})}, 
\end{equation}
where $\bm{A}$ is the matrix extension of  $\alpha_{nc}$ (i.e., $\bm{A} \in 
\mathbb{R}^{N \times C}$).

We impose no further restriction to the bias term $b_c$, so a normal 
zero-mean Gaussian prior is proper
\begin{equation}\label{bias_gaussian}
p(\bm{b}|\bm{\beta}) = \prod_{c=1}^{C}{\mathcal{N}(b_c|0,{\beta_c}^{-1})},
\end{equation}
where ${\beta}_c$ is the inverse variance and $\bm{\beta}$ the vertical 
stacking vector of ${\beta}_c$'s.

To follow the Bayesian framework and encourage the model sparsity, 
hyper-priors over $\bm{A}$ and $\bm{\beta}$ need be defined. The 
truncated Gaussian  belongs to the exponential family and the 
conjugate distribution of the variance of the truncated Gaussian is the 
Gamma distribution (more details in Appendix 
\ref{conjugation_truncated}). The conjugate prior in this paper is 
introduced for the reason that it is very convenient and belongs to 
analytically favorable class of subjective priors to the exponential 
family. Other priors are also alternative such as objective priors 
and empirical priors \cite{beal_variational_2003}. In the 
optimization procedure of our  experiments, most of the weights will 
converge to zero.

The forms of Gamma prior distributions are presented as 
\begin{align} \label{gamma_prior}
& p(\bm{A}) = \prod_{n=1}^{N}\prod_{n=1}^{C} Gamma({\alpha}%
_{nc}|u_1,v_1),
\end{align}
and 
\begin{align}
& p(\bm{b}) \ = \prod_{n=1}^{C} Gamma({\beta}_{c}|u_2,v_2),
\end{align}
where $(u_1, v_1)$ and $(u_2, v_2)$ are hyper-parameters of the Gamma hyper-priors. With these assumptions in
place, marginalizing with respect to 
${\alpha}_{nc}$, we get the complete prior over the weight $w_{nc}$

\begin{align}\label{full_prior_w}
\hspace*{-0.2cm} 
& p(w_{nc}|u_1,v_1) =  {\int_{0}^{\infty} p(w_{nc}|{\alpha}_{nc})p({\alpha}%
_{nc}|u_1,v_1)d{\alpha}_{nc}}  \notag\\ 
&= \delta(f_{nc}w_{nc} \! >\! 0) \sqrt{\frac 2 \pi} {\frac {v_1^{u_1} \Gamma(u_1+{%
\frac 1 2})} {\Gamma(u_1)}} \left(v_1 \!+\! \frac {{w_{nc}}^2}2\right)^{-(u_1+ 
\frac 1 2)}.
\end{align}

For the bias $b_c$, similar to $w_{nc}$, its prior distribution is  
\begin{align}
\label{full_prior_b} \notag p(b_c|u_2, v_2) &= {\int_{0}^{\infty}
	p(b_c|{\beta}_c)p({\beta}_c|u_2,v_2)d{\beta}_c} \\ &=  {\frac {v_2^{u_2} \Gamma(u_2+{\frac 1
			2})} {\sqrt{2 \pi}\Gamma(u_2)}} \left(v_2+ \frac {{b_c}^2} 2\right)^{-(u_2+ \frac 1 2)}.
\end{align}

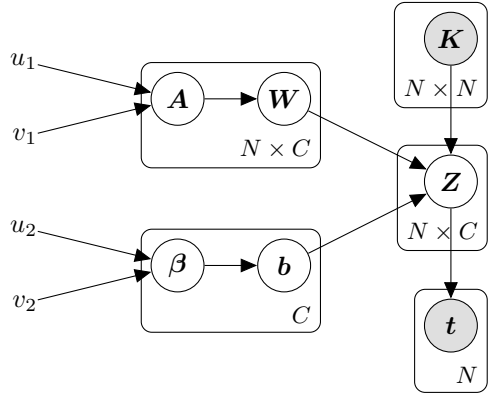
\begin{figure}[ht]
	\begin{center}
		\begin{tabular}{cc}
			
			\begin{tikzpicture}
			\node[latent] (alpha) {$\bm{A}$};
			\node[latent, right=of alpha, xshift=-0.3cm] (w) {$\bm{W}$};
			\node[latent, below=of alpha, yshift=-0.5cm] (beta) {$\bm{\beta}$};
			\node[latent, right=of beta, xshift=-0.3cm] (b) {$\bm{b}$};
			\node[obs, right=of w, yshift=0.8cm, xshift=0.5cm] (x) {$\bm{K}$};
			\node[latent, right=of w, yshift=-1.1cm, xshift=0.5cm] (y) {$\bm{Z}$};
			\node[obs, right=of b, yshift=-0.8cm, xshift=0.5cm] (t) {$\bm{t}$};
			\node[const, left=of alpha, yshift=0.5cm, xshift=-0.5cm] (u1) {$u_1$};
			\node[const, left=of alpha, yshift=-0.5cm, xshift=-0.5cm] (v1) {$v_1$};
			\node[const, left=of beta, yshift=0.5cm, xshift=-0.5cm] (u2) {$u_2$};
			\node[const, left=of beta, yshift=-0.5cm, xshift=-0.5cm] (v2) {$v_2$};			
			\edge {u1, v1} {alpha};
			\edge {u2, v2} {beta};
			\edge {alpha} {w};
			\edge {beta} {b};
			\edge {w, b, x} {y};
			\edge {y} {t};
%
			\plate[] {} {(alpha)(w)} {$N \times C$};
			\plate {} {(beta)(b)} {$C$};
			\plate {} {(x)} {$N\times N$};
			\plate {} {(y)} {$N \times C$};
			\plate {} {(t)} {$N$};
			\end{tikzpicture}
			
		\end{tabular}
	\end{center}
	\caption{Plate diagram of the model's random variables. $u_1$, $v_1$, $u_2$ 
		and $v_2$ are parameters of Gamma distributions. $\bm{A}$ is the parameter of 
		the truncated Gaussian distribution. $\bm{\beta}$ is the parameter of the 
		Gaussian distribution. $\bm{A}$ and $\bm{W}$ are $N \times C$ matrices. $
		\bm{\beta}$ and $\bm{b}$ are $C \times 1$ vectors. $\bm{K}$ is the $N\times N
		$ kernel matrix.  $\bm{Z}$ is a matrix of size  $N \times C$. $\bm{K}$ and $
		\bm{t}$ are shaded to indicate that they are observable.}
	\label{graphic}
\end{figure}

To make these priors non-informative, we fix $u_1,u_2, v_1$ and $v_2$ to small
values  \cite{tipping2001sparse, jiang2019joint}. The plate graph of our proposed model is 
shown in \cref{graphic}.

\subsection{A top-down algorithm: mPCVM$_1$}

\label{em}

This subsection presents the algorithm mPCVM$_1$ by means of the 
derivation of the expectation-maximization (EM) \cite{Dempster1977} 
algorithm that is a general algorithm for the MAP
estimation in the situation where observations are incomplete. In 
the following part, we detail an expectation (E) step and a 
maximization (M) step of the mPCVM$_1$.

The joint posterior probability for $\bm{W}$ and $\bm{b}$ is first estimated in 
the E-step. The posterior probability is expressed as 
\begin{equation}\label{complete_post}
p(\bm{W}, \bm{b}|  \bm{Z},
\bm{A}, \bm{\beta}) = \frac
{p(\bm{Z}|\bm{W},\bm{b})p(\bm{W}|\bm{A})p(\bm{b}|\bm{\beta})} {p(\bm{Z}|\bm{A},\bm{\beta})}.
\end{equation}

Equivalently, the log-posterior is obtained as 
\begin{align}
\label{log-p}
\hspace*{-0.2cm} 
\begin{split}
 {\rm log} &\  p(\bm{W}, \bm{b}| \bm{Z},
 \bm{A}, \bm{\beta})  \\
 & \propto {\rm log} \ 
p(\bm{Z}|\bm{W},\bm{b}) + {\rm log} \ p(\bm{W}|\bm{A}) + {\rm log} \ p(\bm{b}|\bm{\beta}) \\
& \propto  - \bm{b}^{\rm T}\bm{B} \bm{b}  + {\rm \sum \limits_{c=1}^{C}} \big \{ \bm{z_c}^{\rm T} (\bm{\Phi w_c} + {b}_c\bm{1}) - \bm{w_c} ^{\rm T}  \bm{A}_c \bm{w_c} + \\
 &  \quad   (\bm{\Phi w_c}\! + {b}_c\bm{1})^{\rm T} \bm{z_c} \! -\! (\bm{\Phi w_c} + {b}_c\bm{1})^{\rm T}(\bm{\Phi w_c} + {b}_c\bm{1})  \big\},
\end{split}
\end{align} where $\bm{z}_c$, $\bm{w}_c$ and $\bm{\alpha}_c$ are the $c$-th columns of $\bm{Z}$, $\bm{W}$ and $\bm{A}$, respectively, and $\bm{A}_c$ = diag($\bm{\alpha}_c$), $\bm{B}$ = diag($\bm{\beta}$)  diagonal matrices.

Note that a normal zero-mean Gaussian prior over $\bm{W}$ is used in \cref{log-p} for simplicity of computation.
The multi-class classification principle for the mPCVM is ensured in \cref{alpha_nc}

\subsubsection{Expectation Step}

\label{estep}

In the E-step, we should compute the expectation of the log-posterior, 
i.e., Q function
\begin{equation*}
\label{q_func} Q(\bm{W},\bm{b}|\bm{W}^{old},\bm{b}^{old})\newline
\triangleq \mathbb{E}_{\bm{Z},\bm{A},\bm{\beta}}\left[\log \left(p(\bm{W},%
\bm{b}|\bm{Z},\bm{A},\bm{\beta})\right)\right].
\end{equation*}

Hence, we can obtain the Q function:
\begin{align}
\label{q}
\hspace*{-0.25cm} 
\begin{split}
 Q(&\bm{W},\bm{b}|\bm{W}^{old},\bm{b}^{old}) \\
& = - \bm{b}^{\rm T} \bm{\overline B} \bm{b}  + {\rm \sum \limits_{c=1}^{C}} \big \{ \bm{\overline z_c}^{\rm T} (\bm{\Phi w_c} + {b}_c\bm{1})  - \bm{w_c} ^{\rm T} \bm{\overline A}_c \bm{w_c} +\\
& \quad  (\bm{\Phi w_c} + {b}_c\bm{1})^{\rm T} \bm{\overline z_c}  - (\bm{\Phi w_c} + {b}_c\bm{1})^{\rm T}(\bm{\Phi w_c} + {b}_c\bm{1})  
 \big\},
\end{split}
\end{align}
where $\bm{\overline Z} = \mathbb{E}[\bm{Z}|\bm{t}, \bm{W}^{old}, \bm{b}^{old}]$,  $\bm{\overline A}_c = \mathbb{E}[\bm{A}_c|\bm{t}, \bm{W}^{old}, \bm{b}^{old}] $, and $\bm{\overline B} = \mathbb{E}[\bm{B}|\bm{t}, \bm{W}^{old}, \bm{b}^{old}] $. Note that this formula has ignored those items which are not related to $\bm{W}$ or $\bm{b}$. 

The posterior expectation of $z_{nj}$ for all $j \neq t_n$ (assume $t_n = i$) is obtained as 
\begin{align}\label{z_nj}
\hspace*{-0.23cm} 
\begin{split}
&\overline{z}_{nj}= \int z_{nj} p(\bm{z}_n|t_n,\bm{W}^{old},\bm{b}^{old})d\bm{z}_n \\
=& \int z_{nj} \frac {p(t_n, \bm{z}_n|\bm{W}^{old},\bm{b}^{old})} {p(t_n|\bm{W}^{old},\bm{b}^{old})}d\bm{z}_n\\
=& y_{nj} \! - \! \frac{\mathbb{E}_{\varepsilon_{ni}} \bigg [ \mathcal{ N}(\varepsilon_{ni}|y_{nj}\!-\!y_{ni},1) \!\!\!\!\! \prod \limits_{k \neq i,j} \!\!\!\! \Psi(\varepsilon_{ni} \!+\! y_{ni}\!\! -\! y_{nk}) \bigg ] }{\mathbb{E}_{\varepsilon_{ni}} \bigg [ \prod \limits_{k \neq  i} \Psi(\varepsilon_{ni} + y_{ni} - y_{nk}) \bigg ] },
\end{split}
\end{align}
where $\bm{z}_n$ is the $n$-th row of $\bm{Z}$.
The posterior expectation of $z_{ni}$  is
\begin{equation}\label{z_nc}
\begin{aligned}
\overline{z}_{ni} &= \int 
z_{ni} p(\bm{z}_n|t_n,\bm{W}^{old},\bm{b}^{old})d\bm{z}_n\\
&= \int z_{ni} \frac {p(t_n, \bm{z}_n|\bm{W}^{old},\bm{b}^{old})} {p(t_n|\bm{W}^{old},\bm{b}^{old})}d\bm{z}_n\\
&= y_{ni} + \sum_{j \neq i} (y_{nj} - \overline{z}_{nj}).
\end{aligned}
\end{equation}

We present the details in Appendix \ref{pos_exp}. Combined with %
\cref{full_prior_w}, the posterior expectation of $\alpha_{nc}$ is

\begin{align}
\label{alpha_nc}
\begin{split}
\overline{\alpha}_{nc} & =  \int_0^{+\infty}\alpha_{nc} p(\alpha_{nc}|w_{nc},u_1,v_1)d{\alpha}_{nc} \\
& =  \frac {\int_0^{+\infty} \alpha_{nc} p(w_{nc}|\alpha_{nc}) p(\alpha_{nc}|u_1,v_1) d\alpha_{nc}} {p(w_{nc}|u_1,v_1)} \\
&= \frac {2u_1+1} {w_{nc}^2 + 2v_1},
\end{split}
\end{align} if $f_{nc}w_{nc} > 0$, otherwise $\infty$ for $w_{nc} =0$.

Similarly, combined with the \cref{full_prior_b}, the posterior expectation 
of $\beta_c$ is
\begin{align}\label{beta_c}
\begin{split}
\overline{\beta}_c &= \int_0^{\infty}\beta_c p(\beta_c|b_c, u_2, v_2)d\beta _c \\
&= \frac {\int_0^{\infty} \beta_c p(b_c|\beta_c) p(\beta_c| u_2, v_2) d\beta_c} {p(b_c| u_2, v_2)} \\
& = \frac {2u_2+1} {b_c^2 + 2v_2}.
\end{split}
\end{align}

\subsubsection{Maximization Step} 
\label{mstep} 

In the M-step, we need to compute the partial derivatives with respect to $\bm{w_c}$ and
$\bm{b}$:
\begin{equation}\label{dqdw}
\frac {\partial Q} {\partial \bm{w_c}} =  2 \bm{\Phi}^{\rm T} \bm{\overline z_c} - 2 \bm{\Phi} ^{\rm T} \bm{\Phi} \bm{w_c} - 2 b_c \bm{\Phi} ^ {\rm T} \bm{1} - 2 \bm{\overline{A}_c} \bm{w_c},
\end{equation}
\begin{flalign}
\label{dqdb}
& \quad \quad \frac {\partial Q} {\partial b_c} =  2 \bm{1}^{\rm T} \bm{\overline z_c} - 2 \bm{1} ^{\rm T} \bm{\Phi} \bm{w_c} - 2Nb_c- 2 {\overline \beta_c}{b_c}. & 
\end{flalign}

In spite of the difficulty in solving the joint maximization of Q with 
respect to $\bm{W}$ and $\bm{b}$, the optimal $\bm{W}$ and $\bm{b}$ can be 
derived by setting $\partial Q / \partial \bm{W} = 0$ and $\partial Q / \partial \bm{b} = 0$, respectively:

\begin{align}\label{new_w}
w_c^{\rm new} & = (\bm{\Phi}^{\rm T}\bm{\Phi} + diag(\bm{\overline \alpha_c}))^{-1}(\bm{\Phi}^{\rm T}\bm{\overline  z_c} - b_c\bm{\Phi}^{\rm T}\bm{1}), \\
\label{new_b}
b_c^{\rm new} & = \frac{\bm{1}^{\rm T}\bm{\overline z_c} - \bm{1}^{\rm T}\bm{\Phi w_c} }{N + \beta_c}.
\end{align}

The pseudo code of the mPCVM$_1$ can be summarized in Algorithm 1, where 
a majority of $\bm{W}$ would be pruned in the iterations. Hence, the 
mPCVM$_1$ is regarded as a top-down algorithm.

\begin{algorithm}[thp]
	\caption{mPCVM$_1$}
	\label{alg:solution}
	\begin{algorithmic}
		\STATE {\bfseries Input:} train data $\bm{X}$, class label $\bm{t}$, kernel parameter $\theta$.
		\STATE {\bfseries Output:} mPCVM$_1$ classifier, including  $\bm{W}$ and  $\bm{b}$.
		\STATE
		\STATE Compute kernel matrix $\bm{\Phi}$ by kernel function with kernel parameter, randomly initialize $\bm{\overline Z}, \bm{\overline A}$ and $\bm{\overline \beta}$\;
		\REPEAT \STATE $\backslash \backslash$ M step\;
	    \STATE Update $\bm{W}$ and $\bm{b}$ by \cref{new_w} and \cref{new_b}\;
		\STATE Prune elements of the weight $\bm{W}$ if the corresponding elements of $\bm{A}$ are beyond a given threshold \;
		\STATE $\backslash \backslash$ E step\;
		\STATE Update $\overline{\bm{A}}$ and $\overline{\bm{b}}$ by \cref{alpha_nc} and \cref{beta_c}\;
		\STATE Combine the E-step information and update $\overline{\bm{Z}}$ by \cref{z_nj} and \cref{z_nc}\;
		\UNTIL{convergence}
	\end{algorithmic}
\end{algorithm}

\subsection{A bottom-up algorithm: mPCVM$_2$}

\label{mpcvm2}

A constructive framework for fast marginal likelihood maximization is proposed 
in \cite{tipping2003fast} and is extended to the mRVMs in \cite%
{damoulas_inferring_2008} where each sample $n$ is assumed to have the same 
$\alpha_n$ for all classes. Following this framework, we propose the 
mPCVM$_2$ that ensures the multi-class classification principle as the 
mPCVM$_1$. 
Since a sample that belongs to a special class is unequally contributed in the predication of classes, 
we assume that each sample has distinct $\alpha_{nc}$ about each class.

For conciseness, we only consider the weight $\bm{W}$ and  ignore the bias  
$\bm{b}$ by setting $\bm{b} = \bm{0}$. The posterior of $\bm{W}$ is given as 
\begin{align}\label{wpos}
\hspace*{-0.25cm} 
\begin{split}
p&(\bm{W|Z,A}) \propto p(\bm{Z|W})p(\bm{W|A}) \\
& \propto\!\! \prod \limits_{c=1}^{C} \!\mathcal{N}((\bm{\Phi}^{\!\rm T} \!\bm{\Phi} \! + \!\bm{A_{\!c}})^{\!-1}\! \bm{\Phi}^{\!\rm T} \!\bm{z_{\!c}}, (\bm{\Phi}^{\!\rm T}\! \bm{\Phi} \! + \! \bm{A_{\!c}}\!)^{\!-1}\! ) 
\delta(\!\bm{F_{\!c}w_{\!c}}\! >\! 0),\!
\end{split}
\end{align} where $\bm{F_c} = $ diag$ (f_{1c},f_{2c},...,f_{Nc})$. So the MAP for ${w_{nc}}$ could be estimated as 
\begin{align}
\label{w_pos2}
w_{nc} =\delta({w_{nc} f_{nc} > 0 }) ((\bm{\Phi}^{\rm T} \bm{\Phi} + \bm{A_c})^{-1} \bm{\Phi}^{\rm T} {\bm{z}_c})_n.
\end{align}

In the mPCVM$_2$, firstly we compute the marginal likelihood of $p(\bm{Z}|\bm{A}) = \int p(\bm{Z}|\bm{W})p(\bm{W}|\bm{A})d\bm{W}$ with respect to $\bm{W}$.   
We  formulate the marginal likelihood, or equivalently, its logarithm $\mathcal{L}(\bm{A})$:
\begin{equation}\label{type2}
\begin{aligned}
 \mathcal{L}(\bm{A}) &
= \sum\limits_{c=1}^{C} \log p(\bm{z_c}|\bm{\alpha_c})  \\
&= \sum\limits_{c=1}^{C} \log \int p(\bm{z_c}|\bm{w}_c)p(\bm{w}_c|\bm{\alpha_c})d\bm{w_c}\\
&=\sum\limits_{c=1}^{C}  -\frac 1 2 (N\log 2\pi + \log|\bm{\mathcal{ C}}_c| + \bm{z_c}^{\rm T} \bm{\mathcal{ C}}_c^{-1} \bm{z_c}),
\end{aligned}
\end{equation}
where   $\bm{\mathcal{ C}}_c = \bm{I} + \bm{\Phi} {\rm diag(\bm{\alpha_c})} ^{-1} \bm{\Phi}^{\rm T} $ and $\bm{I}$ is the identity matrix. Note 
that we assume $w_{nc}$ is subject to a truncated Gaussian with a scale $\alpha_{nc}$.

Then following the procedure of \cite{tipping2003fast}, we decompose $\bm{\mathcal{C}}_c $ as 
\begin{align}
\begin{split}
\bm{\mathcal{C}}_c 
&= \alpha_{nc}^{-1}\bm{\phi}(\bm{x}_n)\bm{\phi}(\bm{x}_n)^{\rm T} +  \sum \limits_{i \neq n}^{N}\alpha_{ic}^{-1} \bm{\phi}(\bm{x}_i)\bm{\phi}(\bm{x}_i)^{\rm T} \\
& = \alpha_{nc}^{-1}\bm{\phi}(\bm{x}_n)\bm{\phi}(\bm{x}_n)^{\rm T} +  {\bm{\mathcal{C}}_{c}}_{\,-n},
\end{split}
\end{align} where ${\bm{\mathcal{C}}_{c}}_{\, -n}^{\,} $ is ${\bm{\mathcal{C}}_{c}}$ 
without  the contribution of $\bm{\phi}(\bm{x}_n)$. Furthermore, the 
determinant and the inverse of $\bm{\mathcal{C}}_c $ could be decomposed as 
\begin{align}
|\bm{\mathcal{ C}}_c| & = |{\bm{\mathcal{ C}}_c}_{\,-n}^{\,}||1+\alpha_{nc}^{-1}\bm{\phi}(\bm{x}_n)^{\rm T}{\bm{\mathcal{ C}}_c}_{-n}^{-1}\bm{\phi}(\bm{x}_n)|, \\
{\bm{\mathcal{ C}}_c}^{-1}  &  = \; {\bm{\mathcal{ C}}_c}_{-n}^{-1}  - \frac{{\bm{\mathcal{ C}}_c}_{-n}^{-1}\bm{\phi}(\bm{x}_n)\bm{\phi}(\bm{x}_n)^{\rm T}{\bm{\mathcal{ C}}_c}_{-n}^{-1}}{\alpha_{nc} + {\bm{\phi}(\bm{x}_n)^{\rm T}\bm{\mathcal{ C}}_c}_{-n}^{-1}\bm{\phi}(\bm{x}_n)}.
\end{align}  
As everything is ready, $\mathcal{L}(\bm{A})$ is derived as:
\begin{align}
\begin{split}
& \mathcal{L}(\bm{A}) 
 = \sum \limits_{c=1}^{C} \!-\!\frac{1}{2} \bigg[ N\log (2\pi) + \log |{\bm{\mathcal{ C}}_c}_{-n}| + \bm{z}_c^{\rm T} {\bm{\mathcal{ C}}_c}_{-n}^{-1} \bm{z}_c \\
& \qquad  \quad -\log \alpha_{nc} + \log \big( \alpha_{nc} + \bm{\phi}(\bm{x}_n)^{\rm T} {\bm{\mathcal{ C}}_c}_{-n}^{-1} \bm{\phi}(\bm{x}_n) \big)\\
&\qquad \quad - \frac {\big(\bm{\phi}(\bm{x}_n)^{\rm T} {\bm{\mathcal{ C}}_c}_{-n}^{-1} \bm{z_c}\big)^2}{\alpha_{nc} + \bm{\phi}(\bm{x}_n)^{\rm T} { \bm{\mathcal{ C}}_c}_{-n}^{-1} \bm{\phi}(\bm{x}_n)} \bigg ] \\
&\! = \! \sum \limits_{c=1}^{C}\!\bigg [ \!\mathcal{L}(\!{\bm{A}_{\!c}}_{\;-\!n}\!) \!\! + \!\! \frac{1}{2} \! \Big(\!\!\log(\!\alpha_{\!n\!c}\!) \!\!-\! \log(\!\alpha_{\!n\!c} \!\!+\! s_{\!n\!c}\!) \!+\! \frac{q_{nc}^{2}}{\alpha_{\!n\!c}\!\! +\!s_{\!n\!c}}  \! \Big)\!\bigg] \\
& = \sum \limits_{c=1}^{C}\big[ \mathcal{L}({\bm{A}_c}_{\;-n}) + l(\alpha_{nc})\big],
\end{split}
\end{align} where we have the ``sparsity factor" 
\begin{equation}\label{snc}
s_{nc} = \bm{\phi}(\bm{x}_n)^{\rm T } {\bm{\mathcal{ C}}_c}_{-n}^{-1} \bm{\phi}(\bm{x}_n),
\end{equation}
and the ``quality factor" 
\begin{equation}\label{qnc}
q_{nc} = \bm{\phi}(\bm{x}_n)^{\rm T } {\bm{\mathcal{ C}}_c}_{-n}^{-1} \bm{z}_c.  
\end{equation}

By decomposing $\mathcal{L}(\bm{A})$, we isolate the term 
$l(\alpha_{nc})$ that is the contribution of $\alpha_{nc}$ to this 
marginal likelihood. 
Analytically, $\mathcal{L}(\bm{A})$ has a unique maximum with respect to $\alpha_{nc}$

\begin{equation}\label{qgts}
\begin{cases}
\alpha_{nc} = \frac {s_{nc}^2} {q_{nc}^2 - s_{nc}} & \quad \text{if} \ q_{nc}^2 > s_{nc}\\
\alpha_{nc} = \infty & \quad \text{if} \  q_{nc}^2 \leq s_{nc}\\
\end{cases}.
\end{equation}

The pseudo code of the mPCVM$_2$ can be summarized in Algorithm 2.

\begin{algorithm}[thp]
	\caption{mPCVM$_2$}
	\label{alg_2:solution}
	\begin{algorithmic}
		\STATE {\bfseries Input:} train data $\bm{X}$, class label $\bm{t}$, kernel parameter $\theta$.
		\STATE {\bfseries Output:} mPCVM$_2$ classifier, including  $\bm{W}$.
		\STATE
		\STATE Compute kernel matrix $\bm{\Phi}$ by kernel function with kernel parameter, set $\bm{ A} = \infty$, epoch $ =1$ , randomly initialize $\bm{ Z},$ and $N$ is the number of train data. \;
		\REPEAT
		\item \textbf{for}		 $c = 1$ to  $ C$ \textbf{do}
		\item  \quad Compute $s_{nc}$ for $n \in \{1,2,...,N\}$ by \cref{snc} 
		\item  \quad Compute $q_{nc}$ for $n \in \{1,2,...,N\}$ by \cref{qnc}  
		\item  \quad \textbf{if} epoch = 1 \textbf{then}
		\item  \quad \quad  $n = \underset{n} {argmax} \{q_{nc}^2 - s_{nc}  \}$ \;
		\item  \quad \textbf{else} 
		\item  \quad \quad \textbf{if} $\{n|\alpha_{nc}=\infty, q_{nc}^2 > s_{nc}\} \neq \varnothing $ \textbf{then}
		\item \quad \quad \quad $n = \underset{n} {argmax} \{q_{nc}^2 - s_{nc} | \alpha_{nc}=\infty \}$
		\item  \quad \quad \textbf{else if} $\{n|\alpha_{nc} < \infty, q_{nc}^2 < s_{nc}\} \neq \varnothing $ \textbf{then}
				\item \quad \quad \quad $n = \underset{n} {argmin} \{q_{nc}^2 - s_{nc} | \alpha_{nc}<\infty \}$
		\item  \quad \quad \textbf{else} 
		\item  \quad \quad 	\quad	randomly choose n from $\{n|\alpha_{nc}<\infty\}$
		\item  \quad \quad \textbf{end if}
		\item  \quad \textbf{end if}
		\item  \quad \textbf{if} $q_{nc} ^2 > s_{nc}$ and $\alpha_{nc} = \infty$ \textbf{then}
		\item  \quad \quad Set $\alpha_{nc}$ by \cref{qgts}
		\item  \quad \textbf{else if} $q_{nc} ^2 > s_{nc}$ and $\alpha_{nc} < \infty$ \textbf{then}
		\item  \quad \quad Recalculate $\alpha_{nc}$ by \cref{qgts}
     	\item  \quad \textbf{else if} $q_{nc} ^2 \leq s_{nc}$ and $\alpha_{nc} < \infty$ \textbf{then}
  		\item  \quad \quad Set $\alpha_{nc} = \infty$  by \cref{qgts}
		\item  \quad \textbf{end if} 
		\item   \textbf{end for} 
		\STATE Update $\bm{W}$  by \cref{w_pos2} 
		\STATE Update $\bm{Z}$  by \cref{z_nj} and \cref{z_nc}
		\STATE epoch = epoch + 1
		\UNTIL convergence
	\end{algorithmic}
\end{algorithm}

Comparing the mPCVM$_2$ with the mPCVM$_1$, there are four main differences. 
Firstly, they possess different object functions. The mPCVM$_1$ maximizes a 
posterior estimation while the mPCVM$_2$ maximizes the marginal 
likelihood. Secondly, they have different initial states. The mPCVM$_1$ 
initially contains all vectors in the model while the mPCVM$_2$ has only 
one vector for each class. Thirdly, they use very different update strategies. In 
each iteration, the mPCVM$_1$ gradually deletes the vectors that are 
related  with large $\alpha$'s, while exterior vectors can be added or 
vectors that already exist can be deleted in the mPCVM$_2$. Finally, 
they treat basis functions differently. The mPCVM$_1$ considers a vector 
only once. So basis functions that have been deleted cannot be added 
anymore. Conversely, the mPCVM$_2$ might add  a basis function that 
has been removed recently. Hence, the number of iterations of the 
mPCVM$_1$ may be comparably less than the mPCVM$_2$. However, the 
distinguishing characteristic that the mPCVM$_2$ can re-add the vectors 
that have been wrongly deleted in the previous iterations, makes it 
more likely to escape some local optima and gets higher accuracy. In 
a word, the mPCVM$_1$ executes in a top-down fashion while the mPCVM$_2$ is 
like a bottom-up one.

\section{Experimental Studies}
 
\label{exp}

\subsection{Synthetic Data Sets} 
\label{synthetic_data_sets}

This subsection presents experimental results of the mPCVM$_1$, the mPCVM$_2$, the 
mRVM$_1$ and the mRVM$_2$ on two synthetic data sets, i.e., 
\emph{Overlap} and \emph{Overclass},  to analyze their differences.

The data set \emph{Overlap} is generated from several 
different 2-dimensional Gaussian distributions. It contains 3 classes but 
exhibits heavy overlaps. 
In this case,  a nonlinear classifier is necessary. We compare four 
algorithms on this data set and mark the resulting class regions with 
different colors. The relevant vectors are marked with circles. 

In Fig. 3, although the mRVM$_1$ spots the leftmost pivotal relevant 
vector that belongs to the blue class and  locates in the narrow gap 
between two clusters in red, this blue key point contributes wrongly 
to the red class instead of the blue class, resulting in an 
erroneous classification boundary. We infer that the violation of 
the multi-class classification principle for the mPCVM proposed in 
\cref{specification} leads to the degradation of the mRVMs although the 
mRVM$_2$ avoids this vulnerability. In contrast, the mPCVMs, which aim 
to ensure this principle,   manage to get correct regions. Meanwhile, 
the leftmost pivotal blue point should  contribute positively to the 
blue class, contribute negatively to the red class and contribute 
nothing to the black class. However, this point is regarded as a 
relevant vector for all classes in the mRVMs. Its influence on the black 
class could be uncontrollable  noise in this case. In other words, a 
point as a relevant vector should be only for one or two rather than 
all classes, which is apparent in a case with a large number of 
classes. Finally,   although the mRVMs have less relevant vectors 
(the mRVM$_1$   9, the mRVM$_2$ 7) than the mPCVMs (the mPCVM$_1$ 21, the mPCVM$_2$ 7),  in 
the view of  non-zero weights, the mPCVMs  (the mPCVM$_1$ 21, the mPCVM$_2$ 7)  are 
sparser than the mRVMs (the mRVM$_1$ 27, the mRVM$_2$ 21). We argue that the 
mPCVMs are superior to the mRVMs for the reason that the mPCVMs choose precisely 
relevant vectors for each class and impede unnecessary noise.

\begin{figure*}[htp]
	\label{overlap}
	\centering
	\subfigure[mPCVM$_1$]{\includegraphics[width=0.49\linewidth]{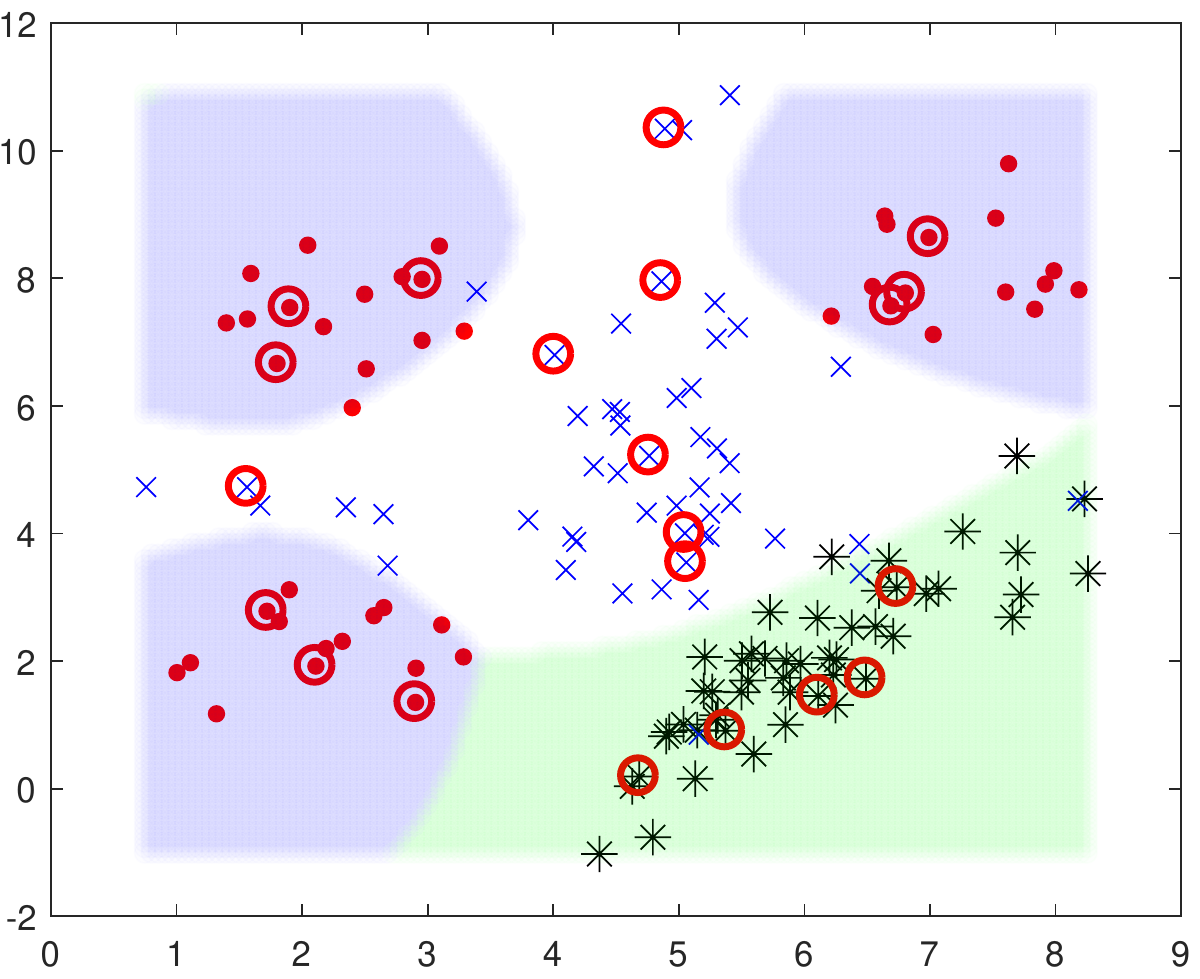} }
	\subfigure[mPCVM$_2$]{\includegraphics[width=0.49\linewidth]{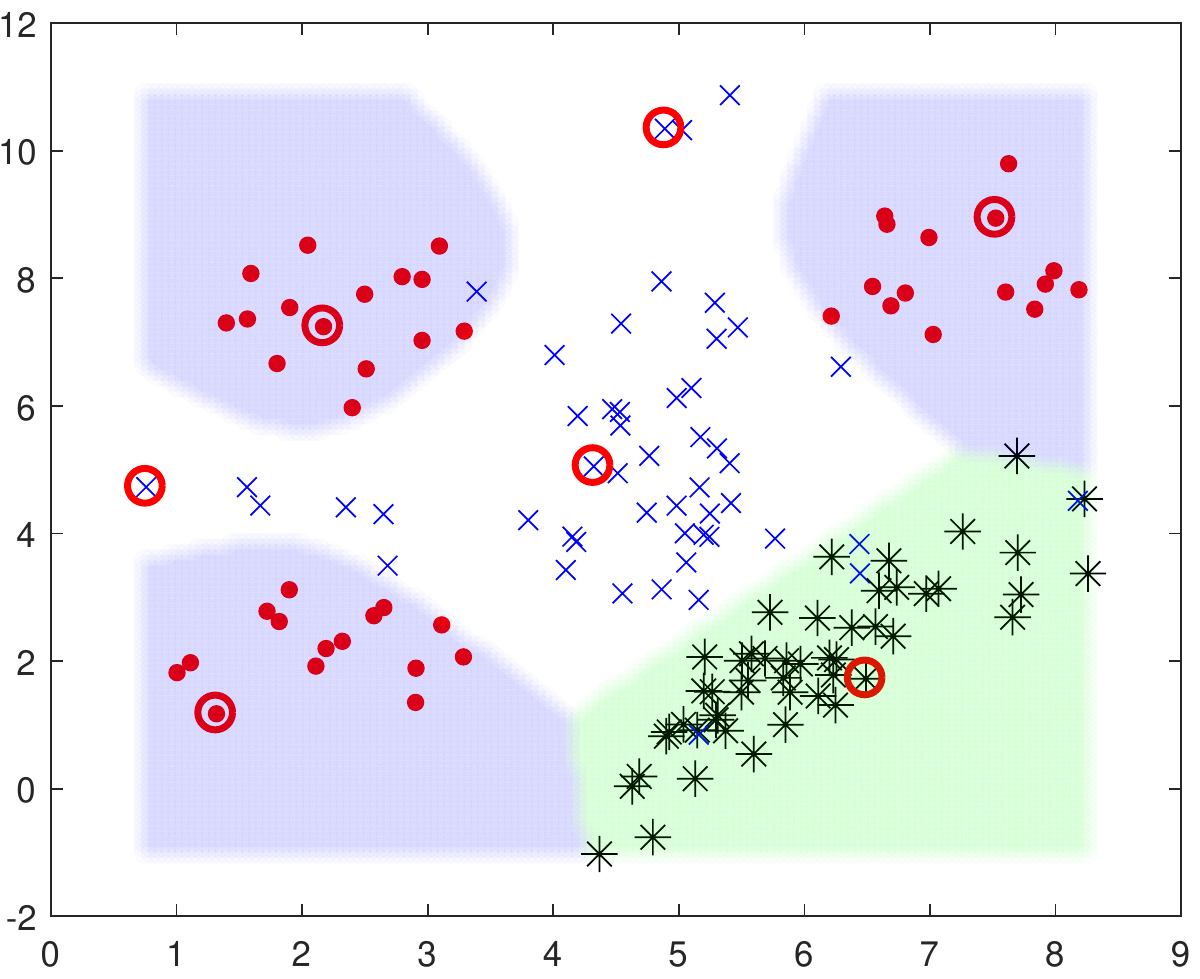}}
	\subfigure[mRVM$_1$]{\includegraphics[width=0.49\linewidth]{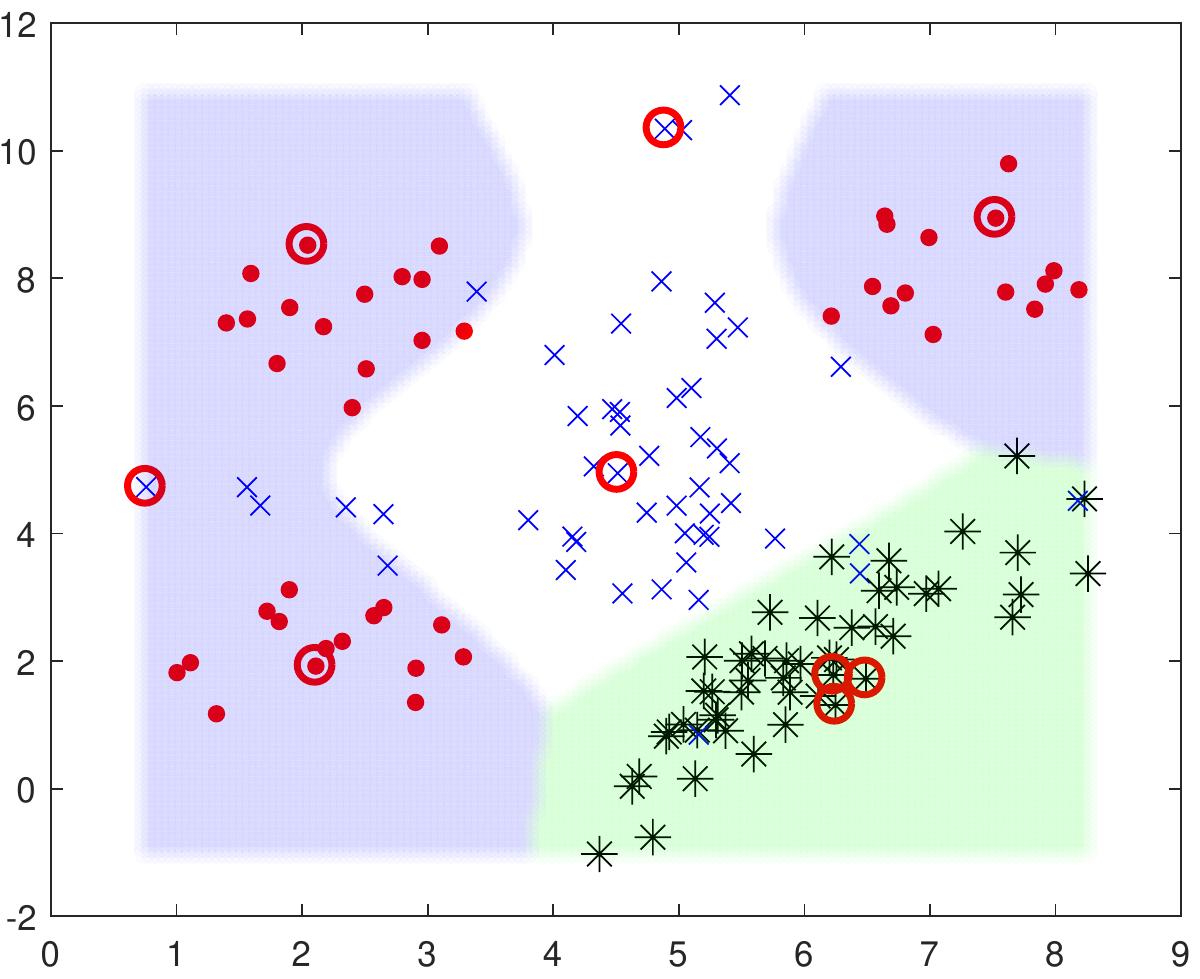}}
	\subfigure[mRVM$_2$]{\includegraphics[width=0.49\linewidth]{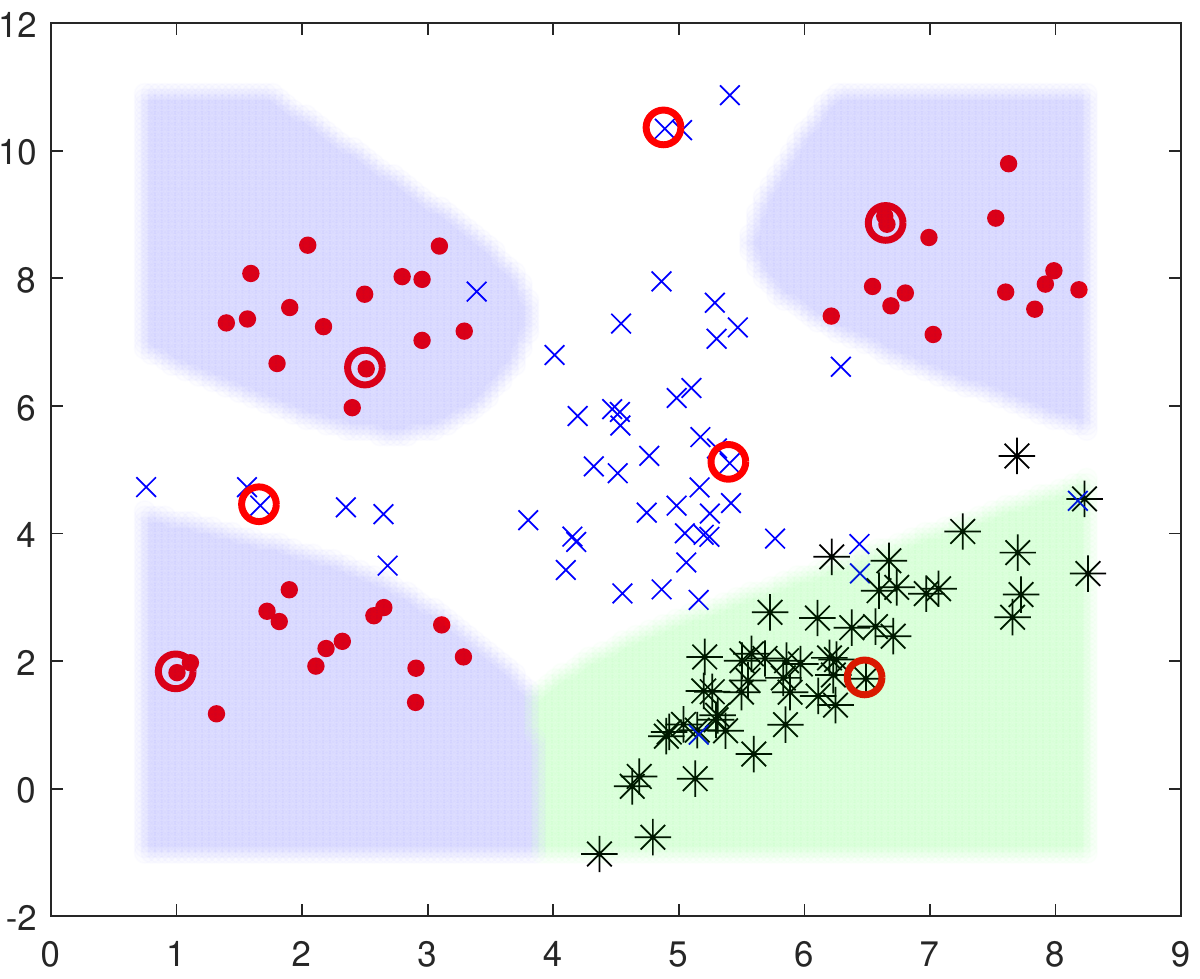}}
	\caption{ 
		The panels illustrate the areas assigned by classifiers to each 
		class. Black points concentrate in the bottom right corner with a skew 
		ellipse shape. Red points appear in three groups split by blue points in the 
		middle canvas. Red circles indicate relevant vectors. In the results, 
		classifiers claim that any point in the green region should be reckoned as the 
		black class, and that any point in the purple region should be reckoned as the red class, 
		and that any point in the white region should be reckoned as the blue class. Four 
		classifiers are compared with the same kernel RBF that uses the same  parameter $\theta = 1$. 
	}
\end{figure*}

The second synthetic data set, \emph{Overclass}, has 127 points in 
ten classes. The challenge of this data is the high imbalance over 
classes. The minor class (pink) has only 3 points while the major 
class (gray) contains 28 points, over 9 times more than those in the 
minor class.

\begin{table}
	\centering
	\caption{ The number of relevant vectors and the number of non-zero weights 
		in the mRVM$_1$, the mRVM$_2$, the mPCVM$_1$ and the mPCVM$_2$. The mRVM$_1$ has no relevant vector in the class 6 and the class 10, which makes the mRVM$_1$ 
		to blur these two classes.  }
	\begin{tabular}{|c|c|c|c|c|}
		\hline
		Class & mRVM$_1$ & mRVM$_2$ & mPCVM$_1$ & mPCVM$_2$ \\
		\hline
		class 1 & 3(30) & 3(30) & 8(8) & 2(2) \\
		\hline
		class 2 & 3(30) & 2(20) & 8(8) & 2(2) \\
		\hline
		class 3 & 2(20) & 3(30) & 7(7) & 2(2) \\	
		\hline
		class 4 & 5(50) & 2(20) & 8(8) & 1(1) \\
		\hline
		class 5 & 5(50) & 1(10) & 7(7) & 1(1)  \\
		\hline
		class 6 & \textbf{0(0)} & 1(10) & 8(8) & 1(1) \\
		\hline
		class 7 & 4(40) & 1(10) & 7(7) & 1(1)  \\
		\hline
		class 8 & 3(30) & 1(10) & 7(7) & 1(1) \\
		\hline
		class 9 & 3(30) & 1(10) & 8(8) & 1(1)  \\
		\hline
		class 10 &  \textbf{0(0)} & 1(10) & 8(8) & 1(1) \\
		\hline
	\end{tabular}
	\label{vector_distribute}
\end{table}

\begin{figure*}
	\label{overclass}
	\centering
	\subfigure[mPCVM$_1$]{\includegraphics[width=0.49\linewidth]{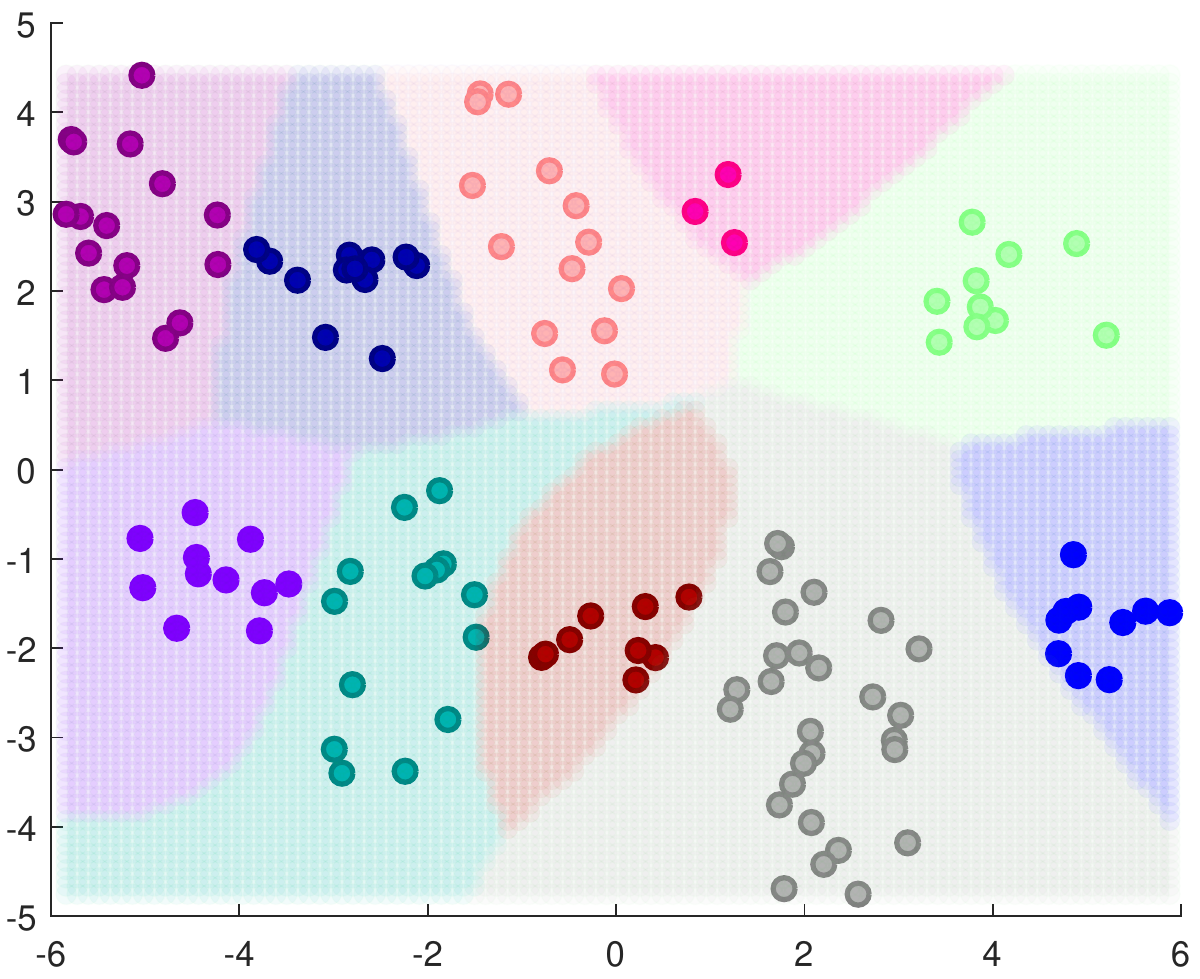}}
	\subfigure[mPCVM$_2$]{\includegraphics[width=0.49\linewidth]{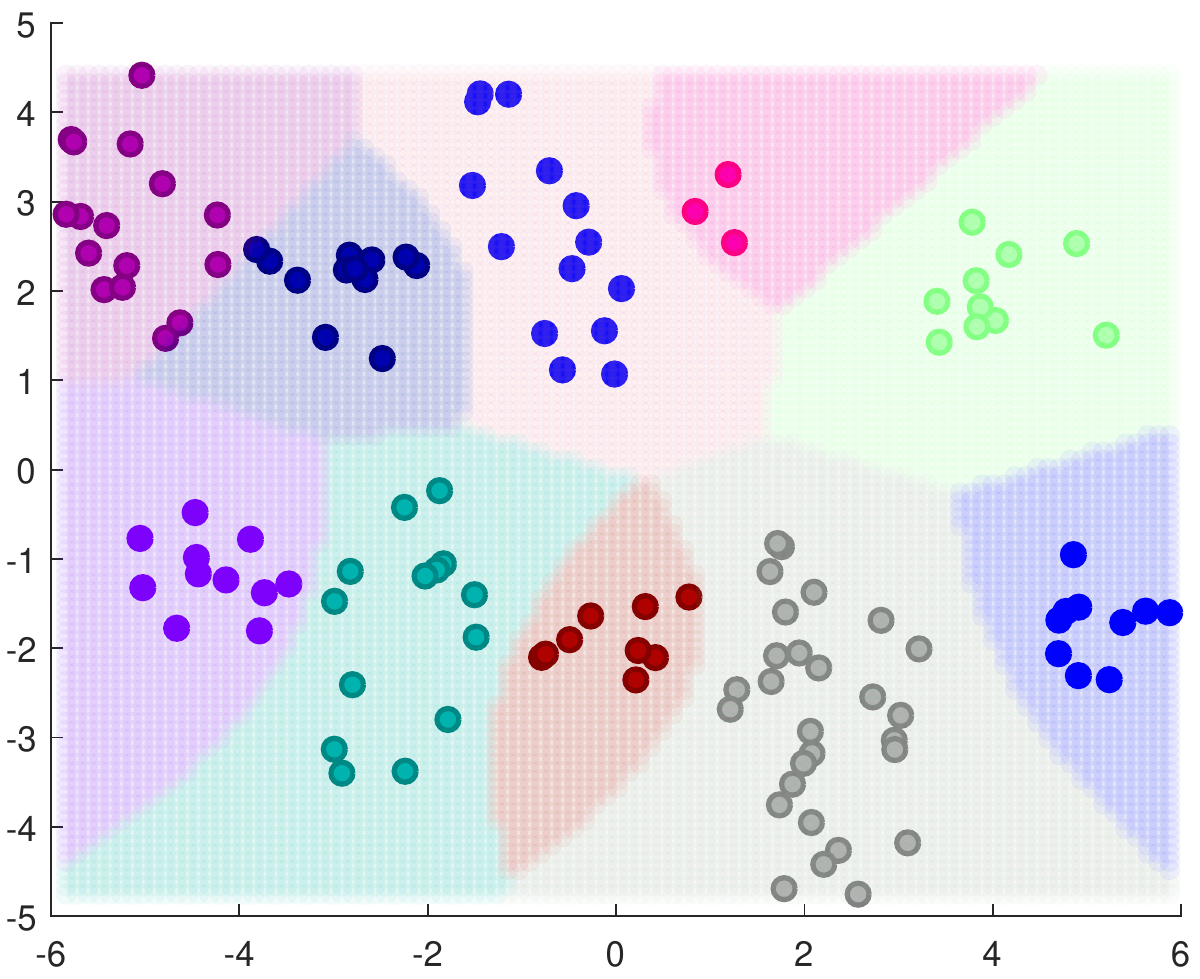}}
	\subfigure[mRVM$_1$]{\includegraphics[width=0.49\linewidth]{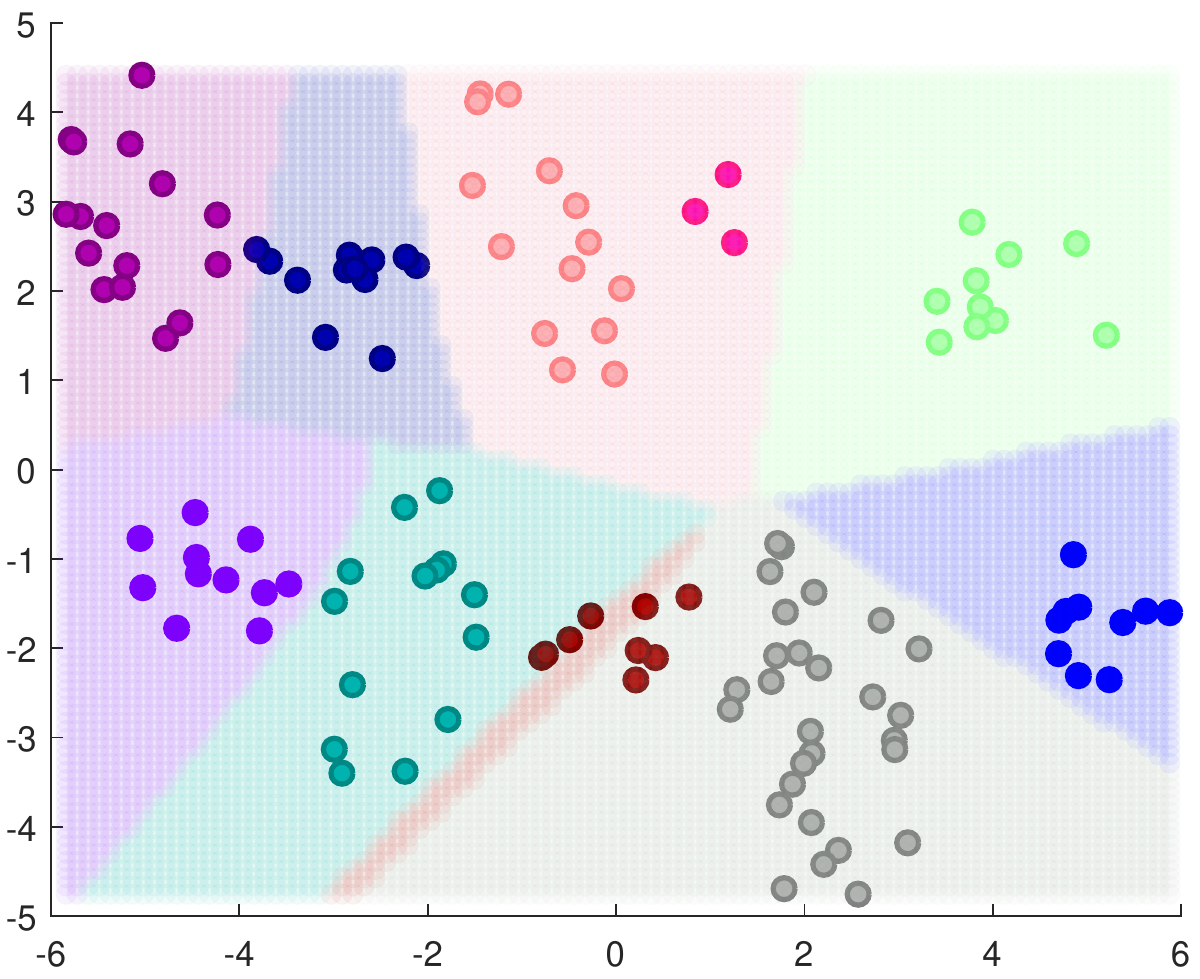}}
	\subfigure[mRVM$_2$]{\includegraphics[width=0.49\linewidth]{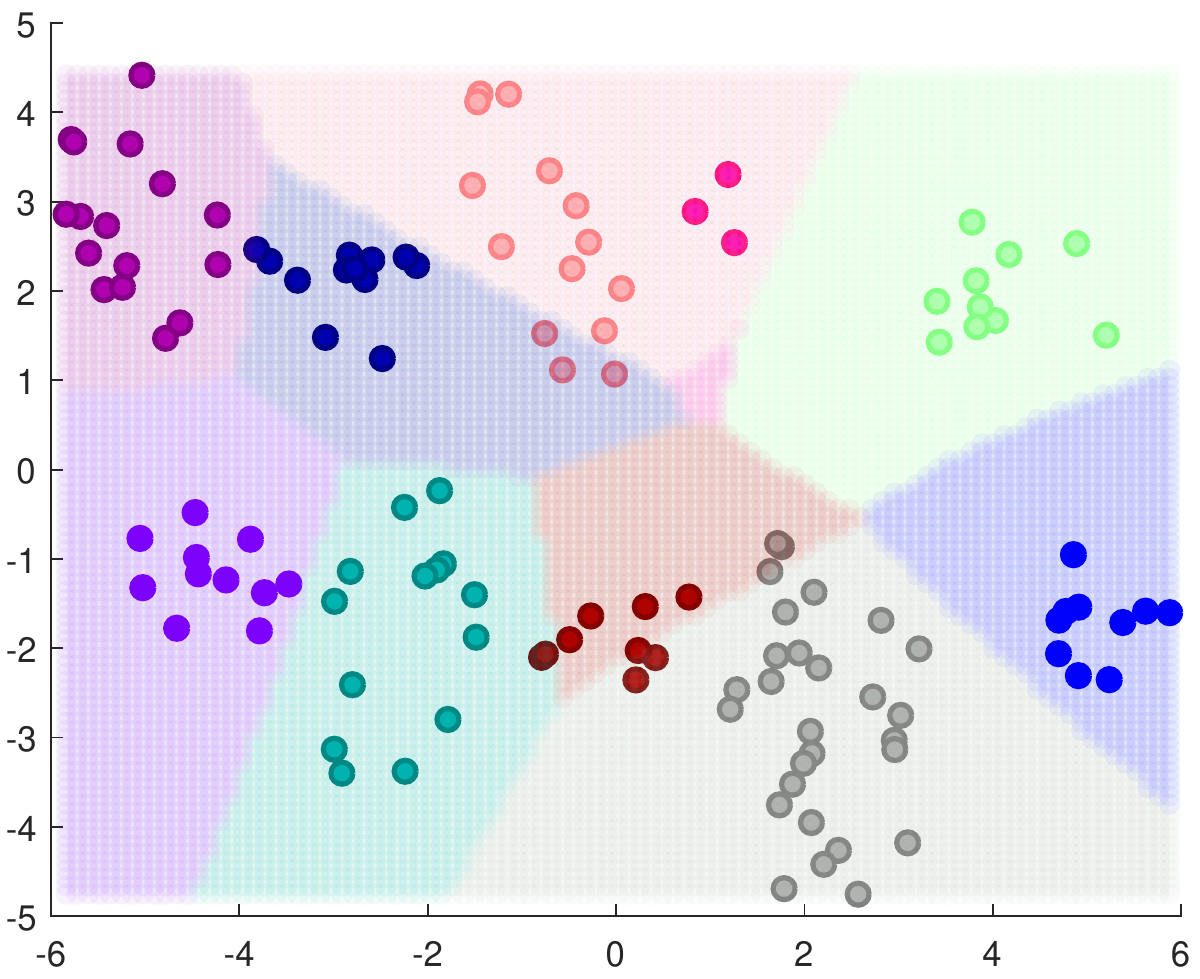}}
	\caption{The patches for classes developed by four classifiers on 
		\emph{Overclass}. Four algorithms  run with the same kernel RBF that uses the same  parameter $\theta = 1$. In this example, the mPCVM$_1$ and the mPCVM$_2$ yield the
		correct results. Yet the mRVM$_1$ and the mRVM$_2$ fail. The mRVMs  ignore wholly
		the class in pink and the mRVM$_1$ generates a narrow band for the class in brown. }
\end{figure*}

We conducted an experiment on \emph{Overclass} with four algorithms. 
In Fig. 4, four algorithms dye every region that is deemed to belong to a 
certain class with a separate color. 
In the current experimental setting, the mRVMs ignore erroneously the minor 
class and consider the pink class to be a part of the crimson class. The area of 
the brown class is narrowed down into a band as its sample size is 
relatively small. 
However, in our algorithms, each class contributes equally to the 
classification boundary. In other words, the classification is 
unlikely to be affected by the sample size of each class. This 
property gives the mPCVMs an advantage when they are applied in the 
experiment where a large number of classes exists yet the class size is imbalanced.

\cref{vector_distribute} lists the number of relevant vectors and 
the number of non-zero weights  for the four learning algorithms. 
Since the absence of relevant vectors belonging to the pink class 
(class 10) and the brown class (class 6), the mRVMs cannot separate out 
the two classes. The classification boundaries in the mRVM$_1$ and the 
mRVM$_2$  are almost linear while those in the mPCVM$_1$ and the mPCVM$_2$ 
are mostly curved. The results show that the mRVM$_1$, the mRVM$_2$, the 
mPCVM$_1$, and the mPCVM$_2$ have 28, 16, 58, and 13 relevant vectors, 
respectively. Each relevant vector in the mRVMs has 10 non-zero weights 
as there are ten classes. So the mRVM$_1$ and the mRVM$_2$ have 280 and 160 
non-zero weights, respectively, while the mPCVM$_1$ and the mPCVM$_2$ have 
76 and 13 non-zero weights, respectively. Therefore, the mPCVMs are sparser than 
the mRVMs. 

\subsection{Benchmark Data Sets}
\label{benchmark}

\begin{table}[!hbp]
	
	\centering
	\caption{Summary of data sets}
	\begin{tabular}{|c|c|c|c|c|}
		\hline
		Data & No.Train & No.test & Dim & Class \\
		\hline 
		Breast   & 546 & 137 & 9 & 2 \\
		\hline
		Glass  & 171 & 43 & 9 & 6 \\
		\hline
		Heart  & 235 & 59 & 16 & 5 \\
		\hline
		Iris & 120 & 30 & 4 & 3 \\
		\hline
		Vowel  & 792 & 198 & 13 & 11 \\
		\hline
		Wine & 142 & 36 & 13 & 3 \\
		\hline
		Wine (red) & 1279 & 320 & 11 & 6 \\
		\hline
		Wine (white) & 3918 & 980 & 11 & 7 \\
		\hline

	\end{tabular}
	\label{dataset_info}
	
\end{table}

In order to evaluate the performance of the mPCVMs, we compare different 
algorithms on 8 benchmark data sets\footnote{%
https://archive.ics.uci.edu/ml/index.php}. The information of these data 
sets is summarized in \cref{dataset_info}. We partition randomly every data 
set into a training set and a test set if no pre-partition exists. In addition, 
standardization is conducted dimensionwise on the training set and the test 
set. 
The comparison algorithms include the mRVM$_1$ \cite{psorakis_multiclass_2010}, 
the mRVM$_2$ \cite{psorakis_multiclass_2010}, the SVM \cite{Chang2011LIBSVM}, the DLSR 
\cite{xiang2012discriminative} and the MLR.
As the mPCVM$_1$ and the mPCVM$_2$ need to set the RBF kernel parameter  $\theta$, we 
follow the method suggested  
in \cite{Raetsch2001}. The hyper-parameter is tuned in the first five 
partitions and the best one is picked according to the  mean accuracy on the 
five partitions.  Then, the performance is measured on the remaining 45 
partitions.

\begin{table*}[!]
	\caption{Comparison of the mPCVM$_1$, the mPCVM$_2$, the mRVM$_1$, the mRVM$_2$, the SVM and  the 
		DLSR on 8 benchmark data sets, under the metrics of the ERR and the AUC. We carried 
		out 45 runs on each data set, and report the averages as well as their 
		standard deviation. }
	\label{performance}\centering%
	\resizebox {6.5 in }{!}{
		\begin{tabular}{|c|c|c|c|c|c|c|c|c|}    
			\hline
			  ERR   &Breast      &Glass        &Heart        &Iris        &Wine         &Wine (red)    &Wine (white)  &Vowel\\ \hline
			mRVM$_1$  &2.952(1.329)&39.380(7.377)&34.501(5.009)&4.667(3.787)&3.765(3.706)&39.674(2.619)&44.386(1.648)&27.374(4.014)\\ \hline
			mRVM$_2$  &3.001(1.649)&33.902(7.478)&35.104(5.605)&4.815(4.178)&2.963(2.865)&40.333(2.253)&43.000(1.287)&6.779(1.996)\\ \hline
			 SVM      &3.277(1.251)&40.672(6.645)&33.484(5.541)&4.370(3.945)&2.160(2.285)&41.549(2.278)&47.220(1.811)&20.393(2.643) \\ \hline
			  DLSR    &2.741(1.156)&47.493(7.657)&49.002(6.622)&9.259(6.738)&4.259(3.276)&50.188(2.313)&56.805(1.700)&55.578(3.268) \\ \hline
			  MLR     &3.244(1.344)&38.191(6.183)&35.104(5.955)&4.519(7.425)&5.926(3.484)&40.271(2.046)&46.150(1.676)&32.907(2.511) \\ \hline
			mPCVM$_1$ &\textbf{2.725(1.189)}&33.127(8.028)&{33.070(5.194)}&{3.852(3.478)}&2.839(2.546)&40.153(2.444)&42.871(1.580)&6.352(2.270) \\ \hline
			mPCVM$_2$ &\textbf{2.725(1.496)}&\textbf{30.439(6.499)}&\textbf{32.316(4.934)}&\textbf{3.185(3.175)}&\textbf{2.099(2.380)}&\textbf{38.250(3.113)}&\textbf{41.128(1.946)}&\textbf{3.592(1.446)} \\ \hline
 
			\hline
			AUC       &Breast       &Glass        &Heart        &Iris         &Wine         &Wine(red)    &Wine(white)  &Vowel\\ \hline
			mRVM$_1$  &98.355(1.093)&82.302(4.815)&82.126(4.538)&99.702(0.532)&99.717(0.505)&76.662(1.774)&72.666(1.126)&97.271(0.537)  \\ \hline
			mRVM$_2$  &98.571(0.908)&85.166(4.686)&80.369(4.811)&99.682(0.623)&99.784(0.352)&76.563(1.672)&74.883(1.161)&99.574(0.284)  \\ \hline
			SVM   	  &\textbf{99.587(0.315)}&81.726(4.495)&84.431(4.715)&99.695(0.565)&99.936(0.134)&74.688(1.640)&69.311(1.264)&98.207(0.312)  \\ \hline
		    DLSR      &99.586(0.312)&79.888(5.519)&\textbf{84.871(4.432)}&93.760(3.678)&99.604(0.389)&74.825(1.708)&69.548(1.114)&86.986(0.887)  \\ \hline
			MLR    	  &99.559(0.340)&81.604(4.753)&83.517(4.590)&99.014(1.602)&97.405(2.187)&76.242(1.690)&71.778(1.166)&95.886(0.512)  \\ \hline
			mPCVM$_1$ &98.394(0.886)&85.737(5.074)&84.016(4.778)&99.699(0.474)&99.761(0.415)&76.069(1.776)&74.320(1.178)&99.634(0.204)  \\ \hline
			mPCVM$_2$ &98.906(0.942)&\textbf{86.899(4.759)}&84.486(3.609)&\textbf{99.777(0.477)}&\textbf{99.879(0.250)}&\textbf{77.025(2.132)}&\textbf{76.087(2.029)}&\textbf{99.709(0.235)}  \\ \hline

		\end{tabular}}
\end{table*}

In our experiment, we select the ERR and the generalized AUC \cite{hand2001simple} 
as metrics to measure performance of these algorithms. The ERR reports classification 
error rates of algorithms. The generalized AUC complements the ERR in measuring 
the performance of algorithms in imbalanced cases. The overall performance 
of an algorithm is measured using the AUC metric:

\begin{equation}\label{auc_define}
\text{AUC}_{generalized} = \frac 2 {C(C-1)} \sum_{i<j}\hat{A}(i,j),
\end{equation}
where $\hat{A}(i,j) = [\hat{A}(i|j)+\hat{A}(j|i)]/2$ is the measure of 
separability between the $i$-th class and the $j$-th one. $\hat{A}(i|j)$ denotes the 
probability that a sample from the $j$-th class is misclassified into the $i$-th one, 
where $i,j \in \{1,2,...,C\}$.  When $ C $ equals to 2, $\hat{A}(1|2) = \hat{A}(2|1)$, 
the generalized AUC is equivalent to the traditional AUC. Notationally, we 
use the AUC instead of the AUC$_{generalized}$ for succinctness.

\cref{performance} reports performance of these algorithms on 8 
benchmark data sets under the metrics of the ERR and the AUC. From the table, the mPCVMs 
perform well in terms of two metrics. For instance, under the ERR metric, the 
mPCVM$_1$ outperforms the comparison algorithms in 6 out of 8 data sets and comes 
second in two cases. In comparison, the mPCVM$_2$  surpasses all other algorithms 
and achieves the best performance. In all data sets, the mPCVMs perform better 
than the mRVMs. This result is partly due to the remediation  of the shortcoming of 
the mRVMs discussed previously.
The mPCVM$_2$ is consistently better than the mPCVM$_1$, 
which demonstrates the superiority  in the methodology that can dynamically 
add and delete relevant vectors.

\subsection{Statistical Comparisons}

To test the statistical significance of results listed in \cref{performance}, 
we apply the Friedman test under the null hypothesis that there is no significant 
difference among algorithms. The alternative hypothesis states there is a 
statistical difference among the tested algorithms. 

The Friedman test statistics is written as $Q = \frac {12N} {k(k+1)} \left[\sum_j R^2_j - \frac {k(k+1)^2} {4} \right]$ where $k$ is the number of 
algorithms, $N$ the number of data sets, and $R_j$ the average rank of 
the $j$-th algorithm. Then, the \textit{p}-value is given by $P(\mathcal{X}^2_{k-1} \geq Q)$. If the  \textit{p}-value is less than 0.10, the null 
hypothesis of the Friedman test should be rejected, and the alternative 
hypothesis should be accepted, indicating that there exists a statistical difference among these 
algorithms.

\begin{table*}[ht]
	\centering
	\caption{The average ranks of the mPCVM$_1$ and the baseline algorithms}
	\begin{tabular}{|c |c |c |c| c| c| c| c|}
		\hline
		Rank  & mRVM$_1$ & mRVM$_2$ & SVM & DLSR  & MLR  & mPCVM$_1$ \\ \hline
		ERR   & 3.375    & 3.375    & 3.75  & 5.375 & 3.875  & 1.25     \\ \hline
		AUC   & 3.375    & 3    & 3.25  & 4.5 & 4.375  & 2.625     \\ \hline
	\end{tabular}
	\label{tab_1_rank}
\end{table*}

\begin{table*}[ht]
	\centering
	\caption{The average ranks of the mPCVM$_2$ and the baseline algorithms}
	\begin{tabular}{|c |c |c |c| c| c| c| c|}
		\hline
		Rank  & mRVM$_1$ & mRVM$_2$ & SVM	& DLSR  & MLR   &  mPCVM$_2$ \\ \hline
		ERR  & 3.5      & 3.375    	& 3.875  & 5.375   & 3.875  &  1     \\ \hline
		AUC  & 3.5    & 3.25    & 3.375  & 4.5 & 4.5 &  1.875     \\ \hline
	\end{tabular}
	\label{tab_2_rank}
\end{table*}

The average ranks of our proposed algorithms and baseline algorithms are 
summarized in \cref{tab_1_rank} and \cref{tab_2_rank} under the two metrics 
where the mPCVM$_1$ and the mPCVM$_2$ are tested separately.

The statistical tests on the mPCVM$_1$ show that this newly proposed algorithm 
could improve the classification accuracy and
that there is no significant difference with regard to the AUC. Meanwhile, 
the statistical tests on the mPCVM$_2$  show an advantage with regard to the 
ERR and the AUC.

If the Friedman test is rejected under the metrics of the ERR and the AUC, a post-hoc test is 
additionally conducted to qualify the difference between our proposed 
algorithms and baseline algorithms. 

The Bonferroni-Dunn test\cite{demvsar2006statistical} is chosen as the 
post-hoc test to compare all algorithms with a control one (the mPCVM$_1$ or the mPCVM$_2$). 
The difference  between two algorithms is significant  if the resulting 
average ranks differ by at least the critical difference, which is written 
as: 
\begin{equation}
CD = q_\alpha \sqrt{\frac{k(k+1)}{6N}},
\end{equation}
where $q_\alpha = 2.326 $ when the significant level $\alpha$ is set as 
0.10 for 6 algorithms. The difference between the $i$-th algorithm and the 
$j$-th one is given by
\begin{equation}
 (R_j - R_i) / \sqrt{\frac {k(k+1)} {6N}}.
\end{equation}

\cref{tab_2_ret} lists Bonferroni-Dunn test results of the mPCVM$_2$. From the table, the 
differences between the mPCVM$_2$ and all the other algorithms are greater than 
the critical difference under the ERR metric, indicating that the pairwise 
difference is significant. Therefore the mPCVM$_2$ performs significantly better 
than the mRVM$_1$, the mRVM$_2$, the SVM, the MLR and the DLSR. Under the AUC metric, the same 
argument holds for the mPCVM$_2$ and the MLR, the DLSR.
However, the differences of the mPCVM$_2$ from the mRVM$_1$, the mRVM$_2$ and the SVM are 
marginally below the critical difference, 
which fails to support the statistical significance  
when $\alpha$ = 0.10. Similarly, \cref{tab_1_ret} shows that the mPCVM$_1$ performs 
significantly better than the mRVM$_1$, the mRVM$_2$, the SVM, the MLR and the DLSR under the 
ERR metric.

\begin{table*}[ht]
	\centering
	\caption{Friedman and Bonferroni-Dunn test results of the mPCVM$_2$ and the baseline 
		algorithms. The threshold is 0.1 and $q_{0.1} = 2.326$. The significant 
		results are marked by bold font}
	\begin{tabular}{|c|c|c||c|c|c|c|c|c|}
		\hline
		& Friedman Q & Friedman \textit{p}-value & CD$_{0.1}$ & mRVM$_1$ & mRVM$_2$ & SVM & DLSR  & MLR   \\ \hline
		ERR   & 23   & 0.000 & 2.176	& \textbf{2.673}    &  \textbf{2.272}   & \textbf{3.074}  & \textbf{4.811} & \textbf{3.207} \\ \hline
		AUC   & 12.714   & 0.026 & 2.176	& {2.138}    &  {1.871}   & {1.871}  & \textbf{3.074} & \textbf{3.074} \\ \hline
	\end{tabular}
	\label{tab_2_ret}
\end{table*}

\begin{table*}[ht]
	\centering
	\caption{Friedman and Bonferroni-Dunn test results of the mPCVM$_1$ and the baseline 
		algorithms. The threshold is 0.1 and $q_{0.1} = 2.326$. The significant 
		results are marked by bold font}
	\begin{tabular}{|c|c|c||c|c|c|c|c|c|}
		\hline
		& Friedman Q & Friedman p-value & CD$_{0.1}$ & mRVM$_1$ & mRVM$_2$ & SVM & DLSR  & MLR   \\ \hline
		ERR   & 20.143   & 0.001 & 2.176	&  \textbf{2.272}   & \textbf{2.272}  & \textbf{2.673} & \textbf{4.410}& \textbf{2.806} \\ \hline
		AUC   & 6.857   & 0.232 & 	&    &     &   &  &  \\ \hline
	\end{tabular}
	\label{tab_1_ret}
\end{table*}

\subsection{Performance under various Classes}
\label{subsection:increase_class} %

A notable fact is that the mPCVM$_2$ improves a small margin over the runner-up in \emph{Breast} (2 classes)
and the improvement in \emph{Vowel} (11 classes) is from 93.221\% 
to 96.408\%. It is not surprising as the mPCVM$_2$ tends to promote better 
accuracy in the data set with a larger number of classes.

To demonstrate the classification performance when the number of classes 
grows, we carried out an experiment in a multi-class data set, \emph{Letter 
Recognition} \cite{frey1991letter}, which contains 26 classes corresponding to 26 capital letters. 
In the experiment, we gradually increased the number of classes and recorded 
the classification accuracies of all the tested algorithms. The experiment 
started with only two classes (600 samples from the class A, 600 samples from the class B). The 
samples were randomly permuted and split into a training set (400) and a 
testing set (800). Then in each round, we added samples 
(200 ones for training and 400 ones for testing) from the next class (e.g. the class C), retrained the models and 
recorded their performance. This process terminated when the first 10 classes 
were added, i.e., A--J. For each algorithm 
except MLR, the best setting was reused as mentioned in \cref{benchmark}. 
Each algorithm run for 45 times per round to obtain an averaged accuracy\footnote{%
Since this data set is balanced, there is no need for the AUC.}.

\begin{figure}
	\centering
	\includegraphics[width=0.5\linewidth]{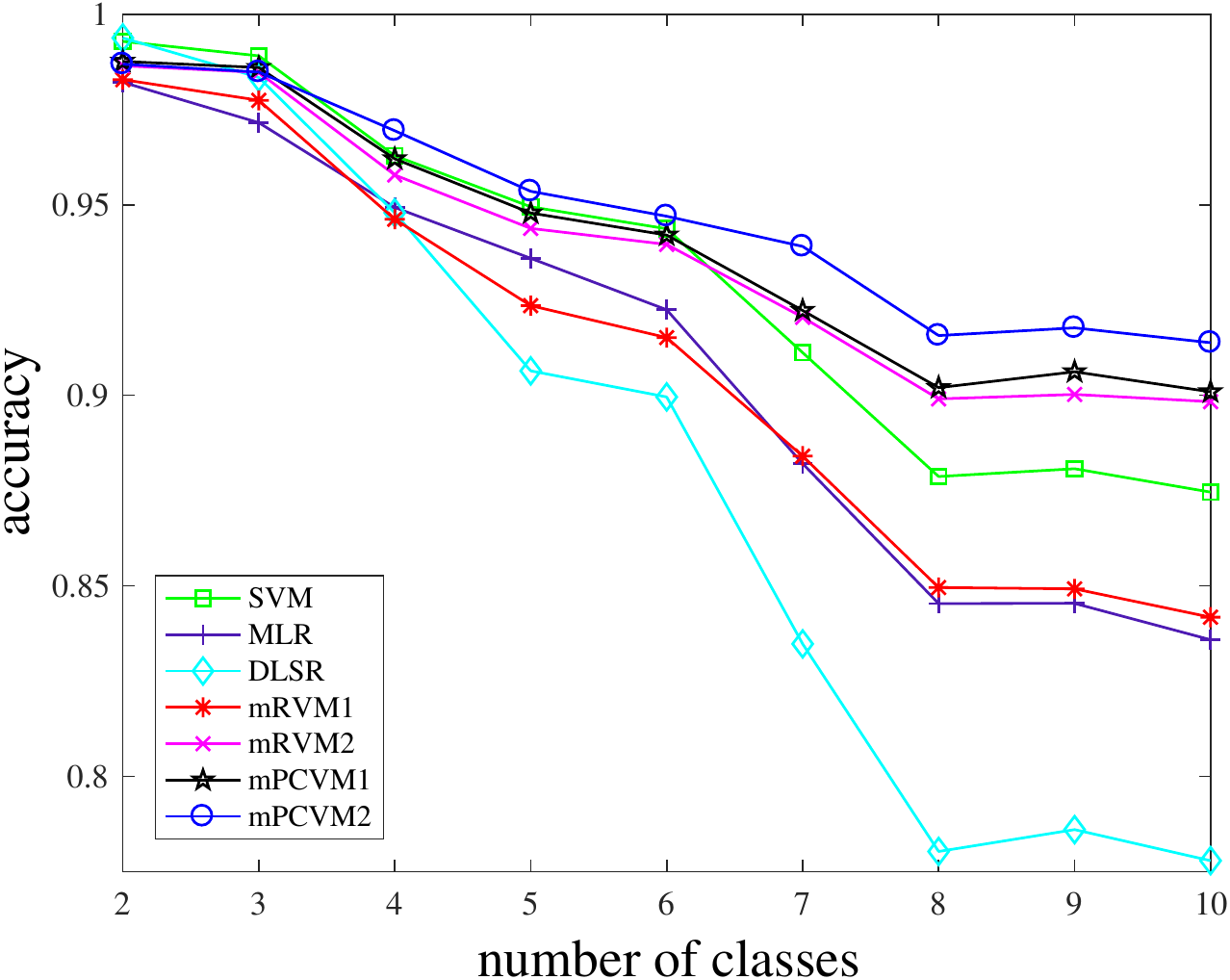}
	\caption{Accuracies of algorithms on \emph{Letter Recognition} with 
		different numbers of classes. X-axis indicates the number of classes and Y-axis 
		presents the accuracies of the algorithms. When the number of classes is small, 
		all the algorithms have similar performance. However, when the number of classes 
		increases, the advantage of our proposed algorithms mPCVMs (esp. the mPCVM$_2$) 
		against the comparison ones has emerged. 
	}
	\label{letter_increase_class}
\end{figure}

The experimental results are illustrated in 
\cref{letter_increase_class}. From the figure, all the  
algorithms have similar results in the binary case. However, as the 
number of classes grows and the problem becomes increasingly 
complicated, the advantage of our proposed algorithms has emerged.  
The mPCVMs (esp. the mPCVM$_2$)  manage to 
offer stable and robust results whilst the performance of other 
algorithms degrades. The mRVMs do not follow the multi-class classification 
principle for the mPCVM, and their performance is unstable and thus loses the 
competition with the mPCVMs. The SVM cannot directly solve multi-class classification 
and this could lead to inferior performance.

\subsection{Experiment on a  data set with a large number of classes} 

Following \cref{subsection:increase_class}, we further increased the number of classes and 
carried out an experiment on \textit{Leaves Plant Species} \cite{mallah2013plant} of 100 classes (labeled from 1 to 100) to evaluate 
the performance of our proposed algorithms and the comparison ones.
In the experiment, all the tested algorithms were trained and evaluated initially on the first 20 classes (1-20) of the data. Then, the number of classes was increased gradually by 20 per round. When the following 20 classes (e.g., 21-40) were added into the training set and the test set of last round, respectively, all the algorithms were retrained and evaluated.  The classification 
accuracies of all the algorithms were recorded in each round. For each algorithm 
except MLR, the best setting was reused as mentioned in \cref{benchmark}. 
Each algorithm run for 45 times per round to obtain an averaged accuracy.

\begin{figure}
	\centering
	\includegraphics[width=0.5\linewidth]{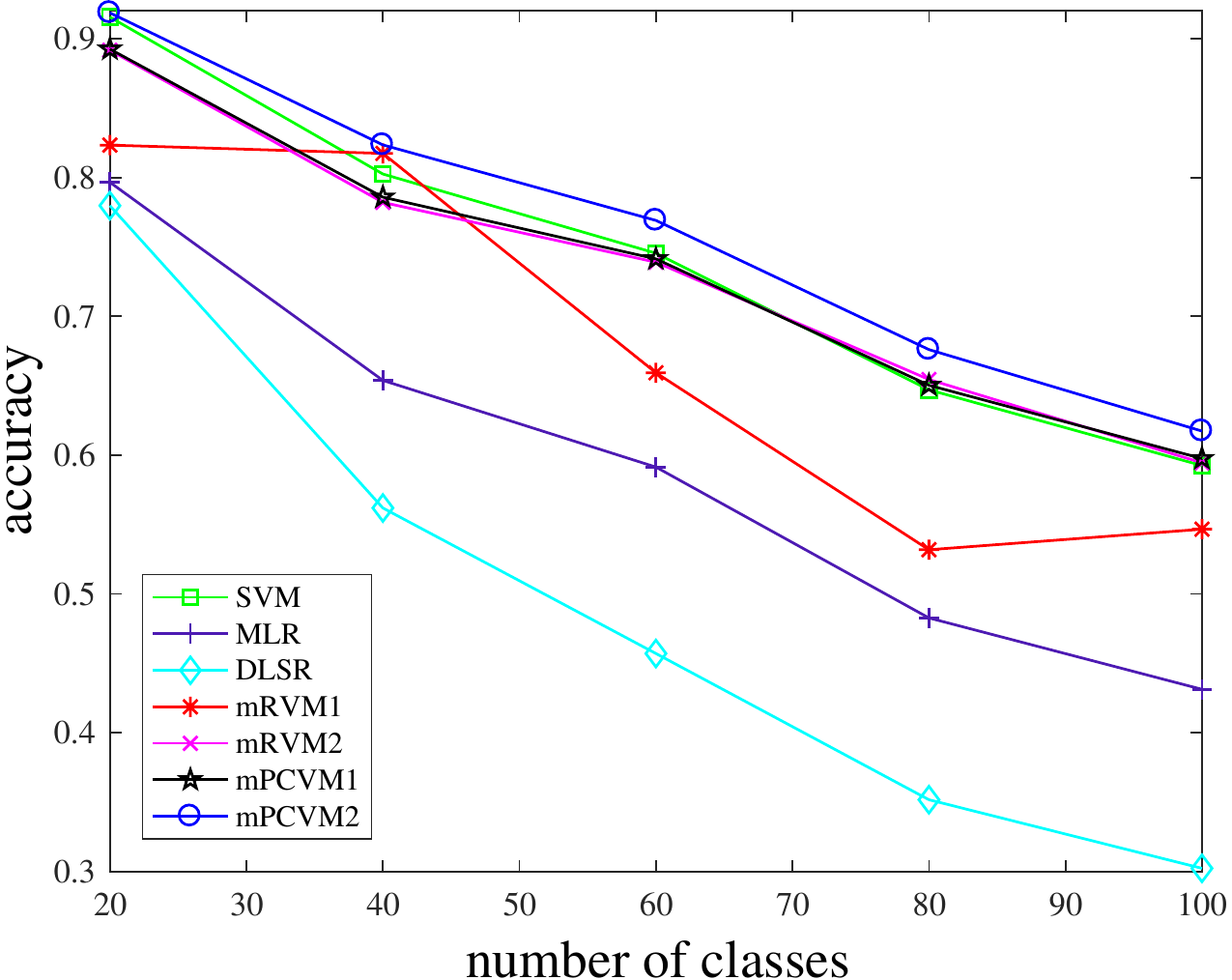}
	\caption{Accuracies of algorithms on \textit{Leaves Plant Species} with 
		different numbers of classes. X-axis indicates the number of classes and Y-axis 
		presents the accuracies of the algorithms. The mPCVM$_1$ obtains similar 
		results to the RVM$_2$ and the SVM. The mPCVM$_2$ surpasses all the comparison algorithms over different class numbers. When the investigated problem contains a large number 
		of classes, the mPCVMs (esp. the mPCVM$_2$) are expected to obtain better performance than the selected benchmark algorithms.
	}
	\label{increase_class_leave}
\end{figure}

The experimental results are illustrated in \cref{increase_class_leave}. The results show
that the mPCVM$_1$ performs similarly to the RVM$_2$ and the SVM, and that the mPCVM$_2$ surpasses
all the comparison algorithms over different class numbers. When the investigated problem contains a large number 
of classes, the mPCVMs (esp. the mPCVM$_2$) are expected to obtain better performance than the selected benchmark algorithms.
The mPCVM$_1$ achieves a comparative performance with others since it ensures the multi-class classification principle.
The mPCVM$_2$ performs consistently better than the mPCVM$_1$ for its flexibility in adding and deleting relevant vectors.

\subsection{Algorithm Complexity}

For the mRVMs and the mPCVMs, they have the same computational complexity   $\mathcal{O}(N^2)$ and the memory 
storage $\mathcal{O}(N^2)$ during computing the kernel matrix,  where $N$ is the number of training samples.
In the following, the computational complexity and the memory storage are considered in the optimization procedure.
Because of the inverse of a kernel matrix that has the shape of $N \times N$, 
the mRVM$_2$ has the computational complexity $\mathcal{O}(N^3)$ and the memory 
storage $\mathcal{O}(N^2)$. 
The mRVM$_1$ has an incremental procedure, so its computational complexity 
decreases to $\mathcal{O}(M^3)$ and its memory storage $\mathcal{O}(MN)$, where 
$M$ is the number of relevant vectors and $M \ll N$. The algorithm mPCVM$_1$ 
initially contains all the $N$ basis functions and then prunes them in each 
class. This could lead to longer training time and larger memory usage. By 
analogy to the mRVM$_2$, its computational complexity is 
$\mathcal{O}(CN^3)$ and its memory storage $\mathcal{O}(CN^2)$ where $C$ is the number of
classes. Similar to the mRVM$_1$, the mPCVM$_2$ is an incremental algorithm, yet it 
has fewer non-zero weights.  Suppose the mPCVM$_2$ has $M_i$ relevant vectors for 
the $i$-th class, its computational complexity is $\mathcal{O}(CM_j^3)$ and its 
memory storage $\mathcal{O}(CM_jN)$  where $M_j= max(M_i) $ for $ i \in 
\{1,2,...,C\}.$

\section{Conclusion and Future Work}

\label{conclude}

In this paper, we propose a multi-class probabilistic classification vector 
machine (mPCVM). This method extends the PCVM into multi-class 
classification. 
We introduce two learning algorithms to optimize the method, a top-down 
algorithm mPCVM$_1$, which is based on an expectation-maximization 
algorithm to obtain the  maximum a \textit{posteriori} point estimates of the parameters,
and a bottom-up algorithm mPCVM$_2$, which is an incremental version of 
maximizing the marginal likelihood. The performance of the mPCVMs is extensively 
evaluated on 8 benchmark data sets. The experimental results conclude that, 
among the selected baseline algorithms, our algorithms have the best 
performance and that the mPCVM$_2$ performs better when the class number is 
large.

However, since the computational complexity is proportional to the number of classes, 
it will be higher than most SVMs.  The relatively higher computational complexity has 
been offset by the superior performance and the benefits to produce the probabilistic 
outputs.  Future work includes reduction of computational complexity with the help of
approximation and combining the idea of the mPCVM with the ensemble learning 
\cite{Chen2009Regularized} .

\section{Appendices}


\subsection{Conjugate Prior of Truncated Gaussian}

\label{conjugation_truncated}

The probability density function of the truncated Gaussian (\cref{truncatd_gaussian}) can be written in the form
\begin{equation*}
\label{exp_truncated}
\hspace*{-0.05cm} 
p(w_{nc}|{\alpha}_{nc}) = \delta(f_{nc}w_{nc}){\left ( \frac {2{\alpha}_{nc}} \pi \right) }^{\frac 1 2}\exp \left (- \frac {{\alpha}_{nc} w_{nc}^2} 2 \right).
\end{equation*}

Comparing with the exponential family
\begin{equation*}
\label{exp_family}
p(\bm{x}|\bm{\eta}) = h(\bm{x})g(\bm{\eta})\exp({\bm{\eta}}^{\rm T}\bm{u}(\bm{x})),
\end{equation*}
we can see that the truncated Gaussian belongs to the exponential family. For any member of the exponential family, there exists a conjugate prior that can be written in the form
\begin{equation*}
\label{exp_family_prior}
p(\bm{\eta}|\bm{a},b) = f(\bm{a},b){g(\bm{\eta})}^b \exp(b {\bm{\eta}}^{\rm T} \bm{a}),
\end{equation*}
where $f(\bm{a},b)$ is a normalization coefficient \cite{BishopPRML}. In this case, we can get
\begin{equation*}
\label{prior_ratio}
p({\alpha}_{nc}|a,b) \propto {\alpha}_{nc}^{\frac b 2} \exp (ab{\alpha}_{nc}).
\end{equation*}

This is exactly $Gamma({\alpha}_{nc}| {\frac b 2} + 1, -ab)$. So the truncated Gaussian has a Gamma distribution as the prior distribution of its precision supposing that $u$ is known (in this case, $u = 0$).

\subsection{Multinomial Probit}
\label{mul_pro}
The multinomial probit is 
\begin{align}
\begin{split}\nonumber
& p(t_n= i|\bm{W},\bm{b}) \\
& = \int \delta (z_{ni} > z_{nj}, \forall j \neq i) \prod\limits_{c=1}^{C} { \mathcal{N}(z_{nc}|y_{nc},1)} d \bm{z_n} \\
& = \int_{- \infty}^{+\infty} \mathcal{N}(z_{ni}|y_{ni},1)  \prod\limits_{j \neq i} \big( \int_{- \infty}^{z_{ni}}{ \mathcal{N}(z_{nj}|y_{nj},1)} d {z_{nj}} \big) d {z_{ni}} \\
& = \int_{- \infty}^{+\infty} \mathcal{N}(z_{ni}|y_{ni},1)\prod\limits_{j \neq i}\Psi(z_{ni} - y_{nj} ) d {z_{ni}} \\
& = \mathbb{E}_{\varepsilon_{ni}} \left[  \prod_{j \neq i}\Psi(\varepsilon_{ni} + y_{ni} - y_{nj}) \right].
\end{split}
\end{align}

\subsection{}
\label{expectation_prove}
$g(\cdot)$ is a differentiable  and bounded function defined in $\mathbb{R}$,  and $\varepsilon$ a random variable, where $\varepsilon \sim \mathcal{N}(0,1)$.
\begin{equation}\nonumber
\begin{aligned}
\mathbb{E}[\varepsilon g(\varepsilon)] &= \frac{1}{\sqrt{2\pi}}\int_{-\infty}^{+\infty} x g(x) e^{-\frac{x^2}{2}} dx \\
&= \frac{1}{\sqrt{2\pi}} \left( \int_{-\infty}^{0} x g(x) e^{-\frac{x^2}{2}} dx + \int_{0}^{+\infty} x g(x) e^{-\frac{x^2}{2}} dx \right) \\
&= \frac{1}{\sqrt{2\pi}} \left( \int^{-\infty}_{0}  g(x) de^{-\frac{x^2}{2}} + \int^{0}_{+\infty}  g(x) de^{-\frac{x^2}{2}} \right) \\
&= \frac{1}{\sqrt{2\pi}} \left( g(x)e^{-\frac {x^2} {2} } |^{-\infty}_{0} - \int_{0}^{-\infty} e^{-\frac {x^2} {2} }  g'(x)dx \right) \\
&+ \frac{1}{\sqrt{2\pi}} \left( g(x)e^{-\frac {x^2} {2} } |_{+\infty}^{0} - \int^{0}_{+\infty} e^{-\frac {x^2} {2} }  g'(x)dx \right).
\end{aligned}
\end{equation}

And $g(\cdot)$ is bounded, so
\begin{equation}\nonumber
\begin{aligned}
\mathbb{E}[\varepsilon g(\varepsilon)] &= \frac{1}{\sqrt{2\pi}} \left (0 - g(0) - \int_{0}^{-\infty} e^{-\frac {x^2} {2} }  g'(x)dx \right) \\
&+ \frac{1}{\sqrt{2\pi}} \left(g(0) - 0 - \int^{0}_{+\infty} e^{-\frac {x^2} {2} }  g'(x)dx \right) \\
&= \frac{1}{\sqrt{2\pi}} \left(\int_{-\infty}^{0}  g'(x) e^{-\frac{x^2}{2}}dx + \int_{0}^{+\infty}  g'(x) e^{-\frac{x^2}{2}}dx \right) \\
&= \mathbb{E}[ g'(\varepsilon)].
\end{aligned}
\end{equation}

In this article, $g(\cdot)$ is $\Psi(\cdot)$. It is bounded by 1 and differentiable.

\subsection{Posterior Expectation}
\label{pos_exp}
The posterior expectation of $z_{nj}$ for all $j \neq t_n$ (assume $t_n = i$) is
\begin{align*}\nonumber
&\overline{z}_{nj} \\
&= \int z_{nj} p(\bm{z}_n|t_n,\bm{W},\bm{b})d\bm{z}_n = \int z_{nj} \frac {p(t_n, \bm{z}_n|\bm{W},\bm{b})} {p(t_n|\bm{W},\bm{b})}d\bm{z}_n\\
&= \frac{1} {p(t_n|\bm{W},\bm{b})} \int_{- \infty}^{+ \infty}\int_{- \infty}^{z_{ni}}  z_{nj} \mathcal{N}(z_{nj}|y_{nj},1)  \\
&  \qquad * \mathcal{N}(z_{ni}|y_{ni},1)\prod 
\limits_{k \neq i,j} \Psi( z_{ni} - y_{nk}) dz_{nj} dz_{ni}   \\
&= \frac{1} {p(t_n|\bm{W},\bm{b})} \int_{- \infty}^{+ \infty}\int_{- \infty}^{z_{ni}}  (z_{nj} - y_{nj} + y_{nj}) \mathcal{N}(z_{nj}|y_{nj},1)  \\
&  \qquad * \mathcal{N}(z_{ni}|y_{ni},1) \prod 
\limits_{k \neq i,j} \Psi( z_{ni} - y_{nk}) dz_{nj} dz_{ni}   \\
&= \frac{1} {p(t_n|\bm{W},\bm{b})} \bigg (y_{nj} \mathbb{E}_{\varepsilon_{ni}} \Big[  \prod_{j \neq i}\Psi(\varepsilon_{ni} + y_{ni} - y_{nj}) \Big] - \\
& \int_{- \infty}^{+ \infty} \mathcal{N}(z_{ni}|y_{ni},1) \mathcal{N}(z_{ni}|y_{nj},1)   \prod 
\limits_{k \neq i,j} \Psi( z_{ni} - y_{nk})dz_{ni}   \bigg) \\
&= y_{nj} - \frac{\mathbb{E}_{\varepsilon_{ni}} \bigg [ \mathcal{ N}(\varepsilon_{ni}|y_{nj}-y_{ni},1) \prod \limits_{k \neq i, j} \Psi(\varepsilon_{ni} + y_{ni} - y_{nk}) \bigg ] }{\mathbb{E}_{\varepsilon_{ni}} \bigg [ \prod \limits_{k \neq  i} \Psi(\varepsilon_{ni} + y_{ni} - y_{nk}) \bigg ] }.
\end{align*} 

And the posterior expectation of $z_{ni}$ (assume $t_n = i$) is 
\begin{align*}\nonumber
&\overline{z}_{ni} = \int z_{ni} p(\bm{z}_n|t_n,\bm{W},\bm{b})d\bm{z}_n \\
&= \int z_{ni} \frac {p(t_n, \bm{z}_n|\bm{W},\bm{b})} {p(t_n|\bm{W},\bm{b})}d\bm{z}_n\\
&= \frac{\int_{- \infty}^{+ \infty}  z_{ni} \mathcal{N}(z_{ni}|y_{ni},1)  \prod 
	\limits_{k \neq i} \Psi( z_{ni} - y_{nk})  dz_{ni}} {p(t_n|\bm{W},\bm{b})}    \\
&= \frac{\int_{- \infty}^{+ \infty}  (\varepsilon_{ni} + y_{ni}) \mathcal{N}(\varepsilon_{ni}|0,1)  \prod 
	\limits_{k \neq i} \Psi(\varepsilon_{ni} + y_{ni} - y_{nk})  d\varepsilon_{ni}} {p(t_n|\bm{W},\bm{b})}    \\
&= y_{ni} + \frac{\mathbb{E}_{\varepsilon_{ni}} \bigg [\varepsilon_{ni} \prod \limits_{k \neq  i} \Psi(\varepsilon_{ni} + y_{ni} - y_{nk}) \bigg ] }{p(t_n|\bm{W},\bm{b})}\\
&= y_{ni} + \frac{ \sum \limits_{j \neq i} \mathbb{E}_{\varepsilon_{ni}} \bigg [\mathcal{N}(\varepsilon_{ni}|y_{nj}\!-y_{ni},1) \!\!\prod  \limits_{k \neq  i,j}\!\! \Psi(\varepsilon_{ni} + y_{ni} - y_{nk}) \bigg ] }{p(t_n|\bm{W},\bm{b})}\\
&= y_{ni} + \sum \limits_{j \neq i}(y_{nj}-\overline{z}_{nj} ).
\end{align*} 

A property of any differentiable  and bounded function is $ \mathbb{E}[{\varepsilon g(\varepsilon) }] = \mathbb{E}[{g{'}(\varepsilon)}]$
 and used in the above step from the 5th line to the 6th line. We prove it in Appendix
\ref{expectation_prove}.

\bibliographystyle{IEEEtran}
\bibliography{tnn}

\end{document}